\documentclass[final,3p,times]{elsarticle}
\usepackage{amsmath}
\usepackage{lineno}
\usepackage{graphicx}    
\usepackage{algorithm}
\usepackage{algpseudocode}
\usepackage{subfigure} 
\usepackage{setspace}  
\usepackage{color}

\newproof{pf}{Proof}
\newtheorem{rmk}{Remark}
\newtheorem{exa}{Example}

\journal{Applied Soft Computing}

\begin{document}

\begin{frontmatter}



\title{EGMM: an Evidential Version of the Gaussian Mixture Model for Clustering}


\author[add1]{Lianmeng Jiao \corref{cor1}}
\ead{jiaolianmeng@nwpu.edu.cn}
\author[add2,add3]{Thierry Den{\oe}ux}
\ead{tdenoeux@hds.utc.fr}
\author[add1]{Zhun-ga Liu}
\ead{liuzhunga@nwpu.edu.cn}
\author[add1]{Quan Pan}
\ead{quanpan@nwpu.edu.cn}

\cortext[cor1]{Corresponding author at: School of Automation, Northwestern Polytechnical
University, Xi'an, 710072, P.R. China. Tel.: +86 29 88431306; Fax: +86 29 88431306.}

\address[add1]{School of Automation, Northwestern Polytechnical University, Xi'an 710072, P.R. China}
\address[add2]{CNRS UMR 7253 Heudiasyc, Universit\'{e} de Technologie de Compi\`{e}gne, Compi\`{e}gne 60200, France}
\address[add3]{Institut Universitaire de France, Paris 75000, France}

\begin{abstract}
The Gaussian mixture model (GMM) provides a simple yet principled framework for clustering, with properties suitable for statistical inference. In this paper, we propose a new model-based clustering algorithm, called EGMM (evidential GMM), in the theoretical framework of belief functions to better characterize cluster-membership uncertainty. With a mass function representing the cluster membership of each object, the evidential Gaussian mixture distribution composed of the components over the powerset of the desired clusters is proposed to model the entire dataset. The parameters in EGMM are estimated by a specially designed Expectation-Maximization (EM) algorithm. A validity index allowing automatic determination of the proper number of clusters is also provided. The proposed EGMM is as simple as the classical GMM, but can generate a more informative evidential partition for the considered dataset. The synthetic and real dataset experiments show that the proposed EGMM performs better than other representative clustering algorithms. Besides, its superiority is also demonstrated by an application to multi-modal brain image segmentation.
\end{abstract}


\end{frontmatter}

\section{Introduction}
%
%
%
%

Clustering is one of the most fundamental tasks in data mining and machine learning. It aims to divide a set of objects into homogeneous groups, by maximizing the similarity between objects in the same group and minimizing the similarity between objects in different groups \cite{Aggarwal14}. As an active research topic, new approaches are constantly proposed, because the usage and interpretation of clustering depend on each particular application. Clustering is currently applied in a variety of fields, such as computer vision \cite{Oskouei21}, security \cite{Chen20}, commerce \cite{Sun21,LiY21}, and biology \cite{Li20b}. Based on the properties of clusters generated, clustering techniques can be classified as partitional clustering and hierarchical clustering \cite{Han11}. Partitional clustering conducts one-level partitioning on datasets. In contrast, hierarchical clustering conducts multi-level partitioning on datasets, in agglomerative or divisive way.

Model-based clustering is a classical and powerful approach for partitional clustering. It attempts to optimize the fit between the observed data and some mathematical model using a probabilistic approach, with the assumption that the data are generated by a mixture of underlying probability distributions. Many mixture models can be adopted to represent the data, among which the Gaussian mixture model (GMM) is by far the most commonly used representation \cite{Melnykov10}. As a model-based clustering approach, the GMM provides a principled statistical way to the practical issues that arise in clustering, e.g., how many clusters there are. Besides, the statistical properties also make it suitable for inference \cite{Fraley02}. The GMM has shown promising results in many clustering applications, ranging from image registration \cite{Ma17}, topic modeling \cite{Costa19}, traffic prediction \cite{Jia19} to anomaly detection \cite{Li20}.

However, the GMM is limited to probabilistic (or fuzzy) partitions for datasets: it does not allow ambiguity or imprecision in the assignment of objects to clusters. Actually, in many applications, it is more reasonable to assign those objects in overlapping regions to a set of clusters rather than some single cluster. Recently, the notion of evidential partition \cite{Denoeux04,Masson08,Denoeux16b} was introduced based on the theory of belief functions \cite{Dempster67,Shafer76,Denoeux17b,Denoeux19b,Denoeux20b}. As a general extension of the probabilistic (or fuzzy), possibilistic, and rough partitions, it allows the object not only to belong to single clusters, but also to belong to any subsets of the frame of discernment that describes the possible clusters \cite{Denoeux16}. Therefore, the evidential partition provides more refined partitioning results than the other ones, which makes it very appealing for solving uncertain data clustering problems. Up to now, different evidential clustering algorithms have been proposed to build an evidential partition for object datasets. Most of these algorithms fall in the category of prototype-based clustering, including evidential $c$-means (ECM) \cite{Masson08}, and its variants such as constrained ECM (CECM) \cite{Antoine12}, median ECM (MECM) \cite{Zhou15}, credal $c$-means(CCM) \cite{Liu15b}, transfer ECM (TECM) \cite{Jiao21}, et al. Besides, in \cite{Denoeux15}, a decision-directed clustering procedure, called EK-NNclus, was developed based on the evidential $k$-nearest neighbor rule, and in \cite{Su19,Gong20}, belief-peaks evidential clustering (BPEC) algorithms were developed by fast search and find of density peaks. Although the above mentioned algorithms can generate powerful evidential partitions, they are purely descriptive and unsuitable for statistical inference. A recent work for model-based evidential clustering was proposed in \cite{Denoeux20} by bootstrapping Gaussian mixture models (called bootGMM). This algorithm builds calibrated evidential partitions in an approximate way, but the high computational complexity in the procedures of bootstrapping and calibration limits its application to large datasets.

In this paper, we propose a new model-based evidential clustering algorithm, called EGMM (evidential GMM), by extending the classical GMM in the belief function framework directly. Unlike the GMM, the EGMM associates a distribution not only to each single cluster, but also to sets of clusters. Specifically, with a mass function representing the cluster membership of each object, the evidential Gaussian mixture distribution composed of the components over the powerset of the desired clusters is proposed to model the entire dataset. After that, the maximum likelihood solution of the EGMM is derived via a specially designed Expectation-Maximization (EM) algorithm. With the estimated parameters, the clustering is performed by calculating the $N$-tuple evidential membership $\mathbf{M} = \{\mathbf{m}_1, \ldots, \mathbf{m}_N\}$, which provides an evidential partition for the considered $N$ objects. Besides, in order to determine the number of clusters automatically, an evidential Bayesian inference criterion (EBIC) is also presented as the validity index. The proposed EGMM is as simple as the classical GMM that has no open parameter and does not require to fix the number of clusters in advance. More importantly, the proposed EGMM generates an evidential partition, which is more informative than a probabilistic partition.

The rest of this paper is organized as follows. Section \ref{sec2} recalls the necessary preliminaries about the theory of belief functions and the Gaussian mixture model from which the proposal is derived. Our proposed EGMM is then presented in Section \ref{sec3}. In Section \ref{sec4}, we conduct experiments to evaluate the performance of the proposal using both synthetic and real-world datasets, and an application to multi-modal brain image segmentation is given to illustrate the interest of the proposal. Finally, Section \ref{sec5} concludes the paper.

\section{Preliminaries} \label{sec2}
We first briefly introduce necessary concepts about belief function theory in Section \ref{sec2.1}. The Gaussian mixture model for clustering is then recalled in Section \ref{sec2.2}.

\subsection{Basics of the Belief Function Theory} \label{sec2.1}
The theory of belief functions \cite{Dempster67,Shafer76}, also known as Dempster-Shafer theory or evidence theory, is a generalization of the probability theory. It offers a well-founded and workable framework to model a large variety of uncertain information. In belief function theory, a problem domain is represented by a finite set $\Omega  = \{\omega_{1}, \ldots, \omega_{C}\}$ called the \textit{frame of discernment}. A \textit{mass function} expressing the belief committed to the elements of ${2^\Omega}$ by a given source of evidence is a mapping function $\mathbf{m}$: ${2^\Omega } \to [0,1]$, such that
\begin{equation}
\mathbf{m}(\emptyset) = 0 \text{~~and~~} \sum\limits_{A \in {2^\Omega }} {\mathbf{m}(A)} = 1.
\end{equation}
Subsets $A \in {2^\Omega }$ such that $\mathbf{m}(A) > 0$ are called the \textit{focal sets} of the mass function $\mathbf{m}$.  The mass function $\mathbf{m}$ has several special cases, which represent different types of information. A mass function is said to be
\begin{itemize}
  \item \textit{Bayesian}, if all of its focal sets are singletons. In this case, the mass function reduces to the precise probability distribution;
  \item \textit{Certain}, if the whole mass is allocated to a unique singleton. This corresponds to a situation of complete knowledge;
  \item \textit{Vacuous}, if the whole mass is allocated to $\Omega$. This situation corresponds to complete ignorance.
\end{itemize}

Shafer \cite{Shafer76} also defined the \emph{belief} and \emph{plausibility} functions as follows
\begin{equation}
{\mathbf{Bel}} (A) = \sum\limits_{B \subseteq A} {\mathbf{m}(B)} {\text{~and~}}{\mathbf{Pl}} (A) = \sum\limits_{B \cap A \ne \emptyset } {\mathbf{m}(B)}, ~~\forall A \in 2^\Omega. \label{eq_BF}
\end{equation}
${\mathbf{Bel}} (A)$ represents the exact support to $A$ and its subsets, and ${\mathbf{Pl}} (A)$ represents the total possible support to $A$. The interval $[{\mathbf{Bel}} (A),{\mathbf{Pl}} (A)]$ can be seen as the lower and upper bounds of support to $A$. The belief functions $\mathbf{m}$, $\mathbf{Bel}$ and  $\mathbf{Pl}$ are in one-to-one correspondence.

For decision-making support, Smets \cite{Smets05} proposed the \emph{pignistic probability} ${\mathbf{BetP}} (A)$ to approximate the unknown probability in $[{\mathbf{Bel}} (A),{\mathbf{Pl}} (A)]$ as follows
\begin{equation}
{\mathbf{BetP}} (A) = \sum\limits_{ A \in B } {\frac{ { \mathbf{m}} (B)}{{\left| B \right|}}},  ~~\forall A \in \Omega, \label{eq_BetP}
\end{equation}
where $\left| B \right|$ is the cardinality of set $B$.

\subsection{Gaussian Mixture Model for Clustering} \label{sec2.2}
Suppose we have a set of objects $\mathbf{X} = \{\mathbf{x}_1, \ldots, \mathbf{x}_N\}$ consisting of $N$ observations of a $D$-dimensional random variable $\mathbf{x}$. The random variable $\mathbf{x}$ is assumed to be distributed according to a mixture of $C$ components (i.e., clusters), with each one represented by a parametric distribution. Then, the entire dataset can be modeled by the following mixture distribution
\begin{equation}
p(\mathbf{x}) = \sum \limits_{k=1} \limits^{C} {\pi_k}{p(\mathbf{x}\mid\mathbf{\theta}_k)},
\end{equation}
where $\mathbf{\theta}_k$ is the set of parameters specifying the $k$th component, and $\pi_k$ is the probability that an observation belongs to the $k$th component ($0 \leq \pi_k \leq 1$, and $\sum_{k=1}^{C}{\pi_k}=1$).

The most commonly used mixture model is the Gaussian mixture model (GMM) \cite{Aggarwal14,Melnykov10,McLachlan19}, where each component is represented by a parametric Gaussian distribution as
\begin{equation}
p(\mathbf{x}\mid\mathbf{\theta}_k) = \mathcal{N}(\mathbf{x}\mid\mathbf{\mu}_k,\mathbf{\Sigma}_k)
= \frac{1}{(2\pi)^{D/2}{|\mathbf{\Sigma}_k|^{1/2}}}\exp\left\{-\frac{1}{2}(\mathbf{x}-\mathbf{\mu}_k)^T {\mathbf{\Sigma}_k}^{-1}(\mathbf{x}-\mathbf{\mu}_k)\right\},
\end{equation}
where $\mathbf{\mu}_k$ is a $D$-dimensional mean vector, $\mathbf{\Sigma}_k$ is a $D \times D$ covariance matrix, and $|\mathbf{\Sigma}_k|$ denotes the determinant of $\mathbf{\Sigma}_k$.

The basic goal of clustering using the GMM is to estimate the unknown parameter $\mathbf{\Theta} = \{\pi_1,\ldots,\pi_C, \mathbf{\mu}_1,\ldots,\mathbf{\mu}_C,$ $\mathbf{\Sigma}_1,\ldots,\mathbf{\Sigma}_C\}$ from the set of observations $\mathbf{X}$. This can be done using maximum likelihood estimation (MLE), with the log-likelihood function given by
\begin{equation}
\log \mathcal{L}(\mathbf{\Theta})=\log p(\mathbf{X}\mid\mathbf{\Theta})=\sum \limits_{i=1} \limits^{N}\log p(\mathbf{x}_i\mid\mathbf{\Theta}) \\
= \sum \limits_{i=1} \limits^{N} \log \sum \limits_{k=1} \limits^{C} {\pi_k} \mathcal{N}(\mathbf{x}_i\mid\mathbf{\mu}_k,\mathbf{\Sigma}_k).
\end{equation}

The above MLE problem is well solved by the Expectation-Maximization (EM) algorithm \cite{Dempster77}, with solutions given by
\begin{equation}
\pi_k^{(s+1)} = \frac{1}{N}\sum \limits_{i=1}\limits^{N}\gamma_{ik}^{(s)}, ~~k=1,\ldots,C,
\end{equation}
\begin{equation}
\mathbf{\mu}_k^{(s+1)} = \frac{\sum_{i=1}^{N}\gamma_{ik}^{(s)}\mathbf{x}_i}{\sum_{i=1}^{N}\gamma_{ik}^{(s)}}, ~~k=1,\ldots,C,
\end{equation}
\begin{equation}
\mathbf{\Sigma}_k^{(s+1)} = \frac{\sum_{i=1}^{N}\gamma_{ik}^{(s)}(\mathbf{x}_i-\mathbf{\mu}_k^{(s)})(\mathbf{x}_i-\mathbf{\mu}_k^{(s)})^T}{\sum_{i=1}^{N}\gamma_{ik}^{(s)}}, ~~k=1,\ldots,C.
\end{equation}
where $\gamma_{ik}^{(s)}$ is the posterior probabilities given the current parameter estimations $\mathbf{\Theta}^{(s)}$ as
\begin{equation}
\gamma_{ik}^{(s)} = \frac{\pi_k^{(s)}\mathcal{N}(\mathbf{x}_i\mid\mathbf{\mu}_k^{(s)},\mathbf{\Sigma}_k^{(s)})}{\sum_{l=1}^{K}\pi_l^{(s)}\mathcal{N}(\mathbf{x}_i\mid\mathbf{\mu}_l^{(s)},\mathbf{\Sigma}_l^{(s)})}, ~~i=1,\ldots,N,~ k=1,\ldots,C.\label{eq_PosP}
\end{equation}
With initialized parameters $\pi_k^{(0)}$, $\mathbf{\mu}_k^{(0)}$, and $\mathbf{\Sigma}_k^{(0)}$, the posterior probabilities and the parameters update alternatively until the change in the log-likelihood becomes smaller than some threshold. Finally, the clustering is performed by calculating the posterior probabilities $\gamma_{ik}$ with the estimated parameters using Eq. (\ref{eq_PosP}).

\section{EGMM: Evidential Gaussian Mixture Model for Clustering} \label{sec3}
Considering the advantages of belief function theory for representing uncertain information, we extend the classical GMM in belief function framework and develop an evidential Gaussian mixture model (EGMM) for clustering. In Section \ref{sec3.1}, the evidential membership is first introduced to represent the cluster membership of each object. Based on this representation, Section \ref{sec3.2} describes how the EGMM is derived in detail. Then, the parameters in EGMM are estimated by a specially designed EM algorithm in Section \ref{sec3.3}. The whole algorithm is summarized and analyzed in Section \ref{sec3.4}. Finally, the determination of the number of clusters is further studied in Section \ref{sec3.5}.

\subsection{Evidential Membership} \label{sec3.1}
Suppose the desired number of clusters is $C$ ($1 < C < N$). The purpose of the EGMM clustering is to assign to the objects in dataset $\mathbf{X}$ soft labels represented by an $N$-tuple evidential membership structure as
\begin{equation}
\mathbf{M} = \{\mathbf{m}_1,  \ldots, \mathbf{m}_N\}, \label{eq_Em}
\end{equation}
where $\mathbf{m}_i$, $i=1,\ldots,N$, are mass functions defined on the frame of discernment $\Omega  = \{\omega_{1}, \ldots, \omega_{C}\}$.

The above evidential membership modeled by mass function $\mathbf{m}_i$ provides a general representation model regarding the cluster membership of object $\mathbf{x}_i$:
\begin{itemize}
  \item When $\mathbf{m}_i$ is a \textit{Bayesian} mass function, the evidential membership reduces to the probabilistic membership of the GMM defined in Eq. (\ref{eq_PosP}) .
  \item When $\mathbf{m}_i$ is a \textit{certain} mass function, the evidential membership reduces to the crisp label employed in many hard clustering methods, such as $c$-means \cite{Jain10}, DPC \cite{Rodriguez14}, et al.
  \item When $\mathbf{m}_i$ is a \textit{vacuous} mass function, the class of object $\mathbf{x}_i$ is completely unknown, which can be seen as an outlier.
\end{itemize}

\begin{exa}\label{ex1}
\begin{spacing}{1.5}
Let us consider a set of $N = 4$ objects $\mathbf{X}=\{\mathbf{x}_1,\mathbf{x}_2,\mathbf{x}_3,\mathbf{x}_4\}$ with evidential membership regarding a set of $C = 3$ classes $\Omega=\{\omega_1,\omega_2,\omega_3\}$. Mass functions for each object are given in Table \ref{tab_membership}. They illustrate various situations: the case of object $\mathbf{x}_1$ corresponds to situation of probabilistic uncertainty ($\mathbf{m}_{1}$ is \textit{Bayesian}); the class of object $\mathbf{x}_2$ is known with precision and certainty ($\mathbf{m}_{2}$ is \textit{certain}), whereas the class of object $\mathbf{x}_3$ is completely unknown ($\mathbf{m}_{3}$ is \textit{vacuous}); finally, the mass function $\mathbf{m}_{4}$ models the general situation where the class of object $\mathbf{x}_4$ is both imprecise and uncertain.
\end{spacing}
\end{exa}

\begin{table}[!ht]
\caption{Example of the evidential membership}
\begin{center}
\begin{tabular}{p{20mm}p{9mm}p{9mm}p{9mm}p{9mm}} 
\hline
$A$                     &$\mathbf{m}_{1}(A)$  &$\mathbf{m}_{2}(A)$ &$\mathbf{m}_{3}(A)$ &$\mathbf{m}_{4}(A)$\\
\hline
$\{\omega_1\}$          &0.2              &0          &0          &0\\
$\{\omega_2\}$          &0.3              &0          &0          &0.1\\
$\{\omega_1,\omega_2\}$ &0                &0          &0          &0\\
$\{\omega_3\}$          &0.5              &1          &0          &0.2\\
$\{\omega_1,\omega_3\}$ &0                &0          &0          &0\\
$\{\omega_2,\omega_3\}$ &0          	  &0          &0          &0.4\\
$\Omega$                &0                &0          &1          &0.3\\
\hline
\end{tabular}
\end{center} \label{tab_membership}
\end{table}

As illustrated in the above example, the evidential membership is a powerful model to represent the imprecise and uncertain information existing in datasets. In the following part, we will study how to derive a soft label represented by the evidential membership for each object in dataset $\mathbf{X}$ given a desired number of clusters $C$.

\subsection{From GMM to EGMM} \label{sec3.2}
In the GMM, each component in the desired cluster set $\Omega  = \{\omega_{1}, \ldots, \omega_{C}\}$ is represented by the following cluster-conditional probability density:
\begin{equation}
p(\mathbf{x}\mid\omega_k, \mathbf{\theta}_k) = \mathcal{N}(\mathbf{x}\mid\mathbf{\mu}_k,\mathbf{\Sigma}_k),~~~ k=1,\ldots,C,
\end{equation}
where $\mathbf{\theta}_k = \{\mathbf{\mu}_k, \mathbf{\Sigma}_k\}$ is the set of parameters specifying the $k$th component $\omega_{k}$, $k=1,\ldots,C$. It means that any objet in set $\mathbf{X} = \{\mathbf{x}_1,  \ldots, \mathbf{x}_N\}$ is drawn from one single cluster in $\Omega$.

Unlike the probabilistic membership in the GMM, the evidential membership introduced in the EGMM enables the object to belong to any subset of $\Omega$, including not only the individual clusters but also the meta-clusters composed of several clusters. In order to model each evidential component $A_j$ ($A_j \neq \emptyset, A_j \in 2^{\Omega} $), we construct the following evidential cluster-conditional probability density:
\begin{equation}
p(\mathbf{x}\mid A_j, \mathbf{\widetilde{\theta}}_j) = \mathcal{N}(\mathbf{x}\mid\mathbf{\widetilde{\mu}}_j,\mathbf{\widetilde{\Sigma}}_j),~~~ j=1,\ldots,M, \label{eq_Ep}
\end{equation}
where $\mathbf{\widetilde{\theta}}_j = \{\mathbf{\widetilde{\mu}}_j, \mathbf{\widetilde{\Sigma}}_j\}$ is the set of parameters specifying the $j$th evidential component $A_j$, $j=1,\ldots,2^{C}-1 \buildrel \Delta \over = M$.

Notice that different evidential components may be nested (e.g., $A_1 = \{\omega_{1}\}$, $A_2 = \{\omega_{1}, \omega_{2}\}$), the cluster-conditional probability densities are no longer independent. To model this correlation, we propose to associate to each component $A_j$ the mean vector $\mathbf{\widetilde{\mu}}_j$ of the average value of the mean vectors associated to the clusters composing $A_j$ as
\begin{equation}
\mathbf{\widetilde{\mu}}_j = \frac{1}{|A_j|} \sum \limits_{k=1} \limits^{C} I_{kj} \mathbf{\mu}_k, ~~ j=1,\ldots,M,\label{eq_ConsMean}
\end{equation}
where $|A_j|$ denotes the cardinal of $A_j$, and $I_{kj}$ is defined as
\begin{equation}
I_{kj}=I(\omega_k \in A_j),~~ k=1,\ldots,C,~~ j=1,\ldots,M.
\end{equation}
with $I(\cdot)$ being the indicator function.

As for the covariance matrix, the values for different components can be free. Some researchers also proposed different assumptions on the component covariance matrix in order to simplify the mixture model \cite{Banfield93}. In this paper, we adopt the following constant covariance matrix:
\begin{equation}
\mathbf{\widetilde{\Sigma}}_j = \mathbf{\Sigma},~~ j=1,\ldots,M, \label{eq_ConsCov}
\end{equation}
where $\mathbf{\Sigma}$ is an unknown symmetric matrix. This assumption results in clusters that have the same geometry but need not be spherical.

In the EGMM, each object is assumed to be distributed according to a mixture of $M$ components over the powerset of the desired cluster set, with each one defined as the evidential cluster-conditional probability density in Eq. (\ref{eq_Ep}). Formally, the evidential Gaussian mixture distribution can be formulated as
\begin{equation}
p(\mathbf{x}) = \sum \limits_{j=1} \limits^{M} {\widetilde{\pi}_j}{p(\mathbf{x}\mid A_j, \mathbf{\widetilde{\theta}}_j)} = \sum \limits_{j=1} \limits^{M} {\widetilde{\pi}_j}{\mathcal{N}(\mathbf{x}\mid \mathbf{\widetilde{\mu}}_j,\mathbf{\widetilde{\Sigma}}_j)},
\end{equation}
where $\widetilde{\pi}_j$ is called mixing probability, denoting the prior probability that the object was generated from $j$th component. Similar with the GMM, the mixing probabilities $\{\widetilde{\pi}_j\}_{j=1}^{M}$ must satisfy $0 \leq \widetilde{\pi}_j \leq 1$, and $\sum_{k=1}^{M}{\widetilde{\pi}_j} = 1$.

\begin{rmk}
\begin{spacing}{1.5}
The above EGMM is an generalization of the classical GMM in the framework of belief functions. When the evidential membership reduces to the probabilistic membership, all the meta-cluster components are assigned zero prior probability, i.e., $\widetilde{\pi}_j = 0$, $j = C+1, \ldots ,M$. In this case, the evidential Gaussian mixture distribution $p(\mathbf{x}) = \sum _{j=1} ^{M} {\widetilde{\pi}_j}{\mathcal{N}(\mathbf{x}\mid \mathbf{\widetilde{\mu}}_j,\mathbf{\widetilde{\Sigma}}_j)} = \sum _{j=1} ^{C} {\widetilde{\pi}_j}{\mathcal{N}(\mathbf{x}\mid \mathbf{{\mu}}_j,\mathbf{{\Sigma}}_j)}$, which is just the classical Gaussian mixture distribution.
\end{spacing}
\end{rmk}

In this formulation of the mixture model, we need to infer a set of parameters from the observations, including the mixing probabilities $\{\widetilde{\pi}_j\}_{j=1}^{M}$ and the parameters for the component distributions $\{\mathbf{\widetilde{\mu}}_j,\mathbf{\widetilde{\Sigma}}_j\}_{j=1}^{M}$. Considering the constraints for the mean vectors and covariance matrices indicated in Eqs. (\ref{eq_ConsMean}) and (\ref{eq_ConsCov}), the overall parameter of the mixture model is $\mathbf{\widetilde{\Theta}} = \{\widetilde{\pi}_1,\ldots,\widetilde{\pi}_M, \mathbf{\mu}_1,\ldots,\mathbf{\mu}_C,$ $\mathbf{\Sigma}\}$. If we assume that the objects in set $\mathbf{X}$ are drawn independently from the mixture distribution, then we can obtain the observed-data log-likelihood of generating all the objects as
\begin{equation}
\log \mathcal{L}_O (\mathbf{\widetilde{\Theta}})=\log p(\mathbf{X}\mid \mathbf{\widetilde{\Theta}})=\sum \limits_{i=1} \limits^{N}\log p(\mathbf{x}_i\mid \mathbf{\widetilde{\Theta}})\\
= \sum \limits_{i=1} \limits^{N} \log \sum \limits_{j=1} \limits^{M} {\widetilde{\pi}_j} \mathcal{N}(\mathbf{x}_i\mid \mathbf{\widetilde{\mu}}_j,\mathbf{\widetilde{\Sigma}}_j).
\label{eq_OL}
\end{equation}
In statistics, maximum likelihood estimation (MLE) is an important statistical approach for parameter estimation. The maximum likelihood estimate of $\mathbf{\widetilde{\Theta}}$ is defined as
\begin{equation}
\mathbf{\widetilde{\Theta}}_{\text{MLE}} = \arg\max \limits_{\mathbf{\widetilde{\Theta}}}{\log \mathcal{L}_O (\mathbf{\widetilde{\Theta}})}, \label{eq_EMLE}
\end{equation}
which is the best estimate in the sense that it maximizes the probability density of generating all the observations. Different from the normal solutions of the GMM, the MLE of the EGMM is rather complicated as additional constraints (see Eqs. (\ref{eq_ConsMean}) and (\ref{eq_ConsCov})) are imposed for the estimated parameters. Next, we will derive the maximum likelihood solution for the EGMM via a specially designed EM algorithm.

\subsection{Maximum Likelihood Estimation via the EM algorithm} \label{sec3.3}
In order to use the EM algorithm to the solve the MLE problem for the EGMM in Eq. (\ref{eq_EMLE}), we artificially introduce a latent variable $\mathbf{z}_i$ to denote the component (cluster) label of each object $\mathbf{x}_i$, $i=1,\ldots,N$, with the form of an $M$-dimensional binary vector $\mathbf{z}_i = (z_{i1},\ldots,z_{iM})$, where,
\begin{equation}
z_{ij} = \left\{ \begin{array}{*{20}{l}}
1, \text{~~if~~} \mathbf{x}_i \text{~belongs to cluster~} j, \\
0, \text{~~otherwise}.\\
\end{array}
\right.
\end{equation}
The latent variable $\mathbf{z}_i$ is independent and identically distributed (idd) according to a multinomial distribution of one draw from $M$ components with mixing probabilities $\widetilde{\pi}_1, \ldots, \widetilde{\pi}_M$. In conjunction with the observed data $\mathbf{x}_i$, the complete data are considered to be $(\mathbf{x}_i,\mathbf{z}_i)$, $i=1,\ldots,N$. Then, the corresponding complete-data log-likelihood can be formulated as
\begin{equation}
\log \mathcal{L}_C (\mathbf{\widetilde{\Theta}})=\log p(\mathbf{X},\mathbf{Z}\mid \mathbf{\widetilde{\Theta}})\\
=\sum \limits_{i=1} \limits^{N} \sum \limits_{j=1} \limits^{M} z_{ij} \log \left[\widetilde{\pi}_j \mathcal{N}(\mathbf{x}_i\mid \mathbf{\widetilde{\mu}}_j,\mathbf{\widetilde{\Sigma}}_j)\right].\label{eq_CL}
\end{equation}
The EM algorithm approaches the problem of maximizing the observed-data log-likelihood $\log \mathcal{L}_O (\mathbf{\widetilde{\Theta}})$ in Eq. (\ref{eq_OL}) by proceeding iteratively with the above complete-data log-likelihood $\log \mathcal{L}_C (\mathbf{\widetilde{\Theta}})$. Each iteration of the algorithm involves two steps called the expectation step (E-step) and the maximization step (M-step). The derivation of the EM solution for the EGMM is detailed in Appendix. Only the main equations are given here without proof.

As the complete-data log-likelihood $\log \mathcal{L}_C (\mathbf{\widetilde{\Theta}})$ depends explicitly on the unobservable data $\mathbf{Z}$, the E-step is performed on the so-called $Q$-function, which is the conditional expectation of $\log \mathcal{L}_C (\mathbf{\widetilde{\Theta}})$ given $\mathbf{X}$, using the current fit for $\widetilde{\Theta}$. More specifically, on the $(s + 1)$th iteration of the EM algorithm, the E-step computes
\begin{equation}
Q(\widetilde{\Theta}, \widetilde{\Theta}^{(s)}) = \mathbb{E}_{\widetilde{\Theta}^{(s)}}[\log \mathcal{L}_C (\mathbf{\widetilde{\Theta}})\mid \mathbf{X}] \\
=\sum \limits_{i=1} \limits^{N} \sum \limits_{j=1} \limits^{M} m_{ij}^{(s)} \log \left[\widetilde{\pi}_j \mathcal{N}(\mathbf{x}_i\mid \mathbf{\widetilde{\mu}}_j,\mathbf{\widetilde{\Sigma}}_j)\right], \label{eq_Qfunc}
\end{equation}
where $m_{ij}^{(s)}$ is the evidential membership of $i$th object to $j$th component, or responsibility of the hidden variable $z_{ij}$, given the current fit of parameters. Using Bayes' theorem, we obtain
\begin{equation}
m_{ij}^{(s)} = p(z_{ij}=1\mid \mathbf{x}_i,\widetilde{\Theta}^{(s)})=\frac{\widetilde{\pi}_j^{(s)} \mathcal{N}(\mathbf{x}_i\mid \mathbf{\widetilde{\mu}}_j^{(s)},\mathbf{\widetilde{\Sigma}}_j^{(s)})}{\sum_{l=1}^{M} \widetilde{\pi}_l^{(s)} \mathcal{N}(\mathbf{x}_i\mid \mathbf{\widetilde{\mu}}_l^{(s)},\mathbf{\widetilde{\Sigma}}_l^{(s)})}, ~~i=1,\ldots,N, ~j=1,\ldots,M. \label{eq_EMemb}
\end{equation}

In the M-step, we need to maximize the $Q$-function $Q(\widetilde{\Theta}, \widetilde{\Theta}^{(s)})$ to update the parameters:
\begin{equation}
\widetilde{\Theta}^{(s+1)} = \arg \max \limits_{\widetilde{\Theta}} Q(\widetilde{\Theta}, \widetilde{\Theta}^{(s)}).
\end{equation}
Different from the observed-data log-likelihood in Eq. (\ref{eq_OL}), the logarithm of $Q$-function in Eq. (\ref{eq_Qfunc}) works directly on the Gaussian distributions. By keeping the evidential membership $m_{ij}^{(s)}$ fixed, we can maximize $Q(\widetilde{\Theta}, \widetilde{\Theta}^{(s)})$ with respect to the involved parameters: the mixing probabilities of the $M$ components $\{\widetilde{\pi}_j\}_{j=1}^{M}$, the mean vectors of the $C$ single-clusters $\{\mu_k\}_{k=1}^{C}$ and the common covariance matrix $\mathbf{\Sigma}$. This leads to closed-form solutions for updating these parameters as follows.
\begin{itemize}
  \item The mixing probabilities of the $M$ components $\widetilde{\pi}_j$, $j= 1,\ldots,M$:
\begin{equation}
\widetilde{\pi}_j^{(s+1)} = \frac{1}{N}\sum \limits_{i=1}\limits^{N}m_{ij}^{(s)}. \label{eq_EMP}
\end{equation}
  \item The mean vectors of the $C$ single-clusters $\mu_k$, $k= 1,\ldots,C$:
\begin{equation}
\mathbf{\Xi}^{(s+1)} = \mathbf{H}^{-1}\mathbf{B}, \label{eq_EMV}
\end{equation}
where $\mathbf{\Xi}^{(s+1)}$ is a matrix of size ($C \times D$) composed of all the mean vectors, i.e., $\mathbf{\Xi}^{(s+1)} = [\mathbf{\mu}_1^{(s+1)};\ldots; \mathbf{\mu}_C^{(s+1)}]$, $\mathbf{H}$ is a matrix of size ($C \times C$) defined by
\begin{equation}
H_{kl} = \sum \limits_{i=1} \limits^{N} \sum \limits_{j=1} \limits^{M} |A_j|^{-2} m_{ij}^{(s)} I_{kj} I_{lj}, ~~k=1,\ldots,C, ~l=1,\ldots,C, \label{eq_H}
\end{equation}
and $\mathbf{B}$ is a matrix of size ($C \times D$) defined by
\begin{equation}
B_{kp} =  \sum \limits_{i=1} \limits^{N} x_{ip} \sum \limits_{j=1} \limits^{M} |A_j|^{-1} m_{ij}^{(s)} I_{kj}, ~~k=1,\ldots,C, ~p=1,\ldots,D. \label{eq_B}
\end{equation}

  \item The common covariance matrix $\mathbf{\Sigma}$:
\begin{equation}
\mathbf{\Sigma}^{(s+1)} = \frac{1}{N} \sum _{i=1}^{N} \sum _{j=1}^{M} m_{ij}^{(s)} (\mathbf{x}_i - \mathbf{\widetilde{\mu}}_j^{(s+1)})(\mathbf{x}_i - \mathbf{\widetilde{\mu}}_j^{(s+1)})^T, \label{eq_ECM}
\end{equation}
where $\mathbf{\widetilde{\mu}}_j^{(s+1)}$ is the updated mean vector of the $j$th evidential component, which is computed using Eq. (\ref{eq_ConsMean}) based on the updated mean vectors of the $C$ single-clusters in Eq. (\ref{eq_EMV}).
\end{itemize}

This algorithm is started by initializing with guesses about the parameters $\widetilde{\Theta}^{(0)}$. Then the two updating steps (i.e., E-step and M-step) alternate until the change in the observed-data log-likelihood $\log \mathcal{L}_O (\mathbf{\widetilde{\Theta}}^{(s+1)}) - \log \mathcal{L}_O (\mathbf{\widetilde{\Theta}}^{(s)})$ falls below some threshold $\varepsilon$. The convergence properties of the EM algorithm are discussed in detail in \cite{McLachlan07}. It is proved that each EM cycle can increase the observed-data log-likelihood, which is guaranteed to convergence to a maximum.

\subsection{Summary and Analysis} \label{sec3.4}
With the estimated parameters via the above EM algorithm, the clustering is performed by calculating the evidential membership $m_{ij}$, $i=1,\ldots,N$, $j=1,\ldots,M$, using Eq. (\ref{eq_EMemb}). The computed $N$-tuple evidential membership $\mathbf{M} = \{\mathbf{m}_1, \ldots, \mathbf{m}_N\}$ provides an evidential partition of the considered objects. As indicated in \cite{Denoeux16}, the evidential partition provides a complex clustering structure, which can boil down to several alternative clustering structures including traditional hard partition, probabilistic (or fuzzy) partition \cite{Bezdek81,DUrso19}, possibilistic partition \cite{Krishnapuram93,Askari17}, and rough partition \cite{Peters15,Namburu17}. We summarize the EGMM for clustering in Algorithm \ref{alg1}.

\begin{algorithm}[ht]
\begin{spacing}{1.5}
 \caption{EGMM clustering algorithm.}
 \label{alg1}
 \begin{algorithmic}[1]
 \Require
  $\mathbf{X} = \{\mathbf{x}_1, \ldots, \mathbf{x}_N\}$: $N$ samples in $\mathbb{R}^{D}$; ~$C$: number of clusters $1 < C<N$; ~$\varepsilon$: termination threshold.
 \State Initialize the mixing probabilities $\{\widetilde{\pi}_j^{(0)}\}_{j=1}^{M}$, the mean vectors of the $C$ single-clusters $\{\mathbf{\mu}_k^{(0)}\}_{k=1}^{C}$ and the common covariance matrix $\mathbf{\Sigma}^{(0)}$;
 \State $s\leftarrow 0$;
 \Repeat
      \State Calculate the mean vectors $\{\mathbf{\widetilde{\mu}}_j^{(s)}\}_{j=1}^{M}$ and the covariance matrices $\{\mathbf{\widetilde{\Sigma}}_j^{(s)}\}_{j=1}^{M}$ of the evidential components using the current parameters $\{\mathbf{\mu}_k^{(s)}\}_{k=1}^{C}$ and $\mathbf{\Sigma}^{(s)}$ based on Eq. (\ref{eq_ConsMean}) and Eq. (\ref{eq_ConsCov}), respectively;
      \State Calculate the evidential memberships $\{\{m_{ij}^{(s)}\}_{i=1}^{N}\}_{j=1}^{M}$ using the current mixing probabilities $\{\widetilde{\pi}_j^{(s)}\}_{j=1}^{M}$, the mean vectors $\{\mathbf{\widetilde{\mu}}_j^{(s)}\}_{j=1}^{M}$ and the covariance matrices $\{\mathbf{\widetilde{\Sigma}}_j^{(s)}\}_{j=1}^{M}$ based on Eq. (\ref{eq_EMemb});
      \State Update the mixing probabilities $\{\widetilde{\pi}_j^{(s+1)}\}_{j=1}^{M}$ and the mean vectors of the $C$ single-clusters $\{\mathbf{\mu}_k^{(s+1)}\}_{k=1}^{C}$ using the calculated evidential memberships $\{\{m_{ij}^{(s)}\}_{i=1}^{N}\}_{j=1}^{M}$ based on Eqs. (\ref{eq_EMP})-(\ref{eq_B});
      \State Update the common covariance matrix $\mathbf{\Sigma}^{(s+1)}$ using the calculated evidential memberships $\{\{m_{ij}^{(s)}\}_{i=1}^{N}\}_{j=1}^{M}$ and the updated mean vectors $\{\mathbf{\mu}_k^{(s+1)}\}_{k=1}^{C}$ based on Eq. (\ref{eq_ECM});
      \State Compute the observed-data log-likelihood $\log \mathcal{L}_O (\mathbf{\widetilde{\Theta}}^{(s+1)})$ based on the updated parameters using Eq. (\ref{eq_OL});
      \State $s \leftarrow s+1$;
 \Until $\log \mathcal{L}_O (\mathbf{\widetilde{\Theta}}^{(s)}) - \log \mathcal{L}_O (\mathbf{\widetilde{\Theta}}^{(s-1)}) < \varepsilon$;\\
\Return the $N$-tuple evidential membership $\mathbf{M} = \{\mathbf{m}_1, \ldots, \mathbf{m}_N\}$, with each $\mathbf{m}_i = \{m_{ij}^{(s)}\}_{j=1}^{M}$, $i=1,\ldots,N$.
 \end{algorithmic}
\end{spacing}
\end{algorithm}

\textbf{Generality Analysis}:
The proposed EGMM algorithm provides a general framework for clustering, which boils down to the classical GMM when we constrain all the evidential components to be singletons, i.e., $A_j = \omega_j$, $j = 1, \ldots, C$. Compared with the GMM algorithm, the evidential one allocates for each object a mass of belief to any subsets of possible clusters, which allows to gain a deeper insight on the data.

\textbf{Convergence Analysis}:
As indicated in \cite{Park09}, the EM algorithm for mixture models takes many iterations to reach convergence, and reaches multiple local maxima starting from different initializations. In order to find a suitable initialization and speed up the convergence for the proposed EGMM algorithm, it is recommended to run the $c$-means algorithm \cite{Jain10} and choose the means of the clusters and the average covariance of the clusters for initializing $\{\mathbf{\mu}_k^{(0)}\}_{k=1}^{C}$ and $\mathbf{\Sigma}^{(0)}$, respectively. As for the mixing probabilities $\{\widetilde{\pi}_j^{(0)}\}_{j=1}^{M}$, if no prior information is available, these values can be initialized equally as $1/M$.

\textbf{Complexity Analysis}:
For each object, the proposed EGMM algorithm distributes a fraction of the unit mass to each non-empty element of $2^{\Omega}$. Consequently, the number of parameters to be estimated is exponential in the number of clusters $C$ and linear in the number of objects $N$. Considering that, in most cases, the objects assigned to elements of high cardinality are of less interpretability, in practice, we can reduce the complexity by constraining the focal sets to be composed of at most two clusters. By this way, the number of parameters to be estimated is drastically reduced from $O(2^{C} N)$ to $O(C^2 N)$.

\subsection{Determining the Number of Clusters} \label{sec3.5}
One important issue arising in clustering is the determination of the number of clusters. This problem is often referred to as cluster validity. Most of the methods for GMM clustering usually start by obtaining a set of partitions for a range of values of $C$ (from $C_{\min}$ to $C_{\max}$) which is assumed to contain the optimal $C$. The number of clusters is then selected according to
\begin{equation}
C^{\ast} = \arg\max \limits_{C_{\min} \leq C \leq C_{\max}} \mathcal{I} (\mathbf{\Theta}_C, C),
\end{equation}
where $\mathbf{\Theta}_C$ is the estimated parameter with $C$ clusters, and $\mathcal{I} (\mathbf{\Theta}_C, C)$ is some validity index. A very common criterion can be expressed in the form \cite{Aggarwal14}
\begin{equation}
\mathcal{I} (\mathbf{\Theta}_C, C) =\log \mathcal{L}(\mathbf{\Theta}_C) - \mathcal{P}(C),
\end{equation}
where $\log \mathcal{L}(\mathbf{\Theta}_C)$ is the maximized mixture log-likelihood when the number of clusters is chosen as $C$ and $\mathcal{P}(C)$ is an increasing function penalizing higher values of $C$.

Many examples of such criterion have been proposed for the GMM, including Bayesian approximation criteria, such as Laplas-empirical criterion (LEC) \cite{McLachlan00}, and Bayesian inference criterion (BIC) \cite{Fraley02}, and information-theoretic criterion, such as minimum description length (MDL) \cite{Grunwald07}, minimum message length (MML) \cite{Yatracos15}, and Akaike's information criterion (AIC) \cite{Charkhi18}. Among these criterion, the BIC has given better results in a wide range of applications of model-based clustering. For general mixture models, the BIC is defined as
\begin{equation}
\text{BIC}(\mathbf{\Theta}_C, C)  = \log \mathcal{L}(\mathbf{\Theta}_C) - \frac{1}{2}v_{C}\log(N),
\end{equation}
where $v_{C}$ is the number of independent parameters to be estimated in $\mathbf{\Theta}_C$ when the number of clusters is chosen as $C$.

For our proposed clustering approach, we adopt the above BIC as the validity index to determine the number of clusters. For EGMM, the mixture log-likelihood is replaced by the evidential Gaussian mixture log-likelihood $\log \mathcal{L}_O (\mathbf{\widetilde{\Theta}}_C)$ defined in Eq. (\ref{eq_OL}), and the number of independent parameters in $\mathbf{\Theta}_C$ is replaced by that in $\mathbf{\widetilde{\Theta}}_C$. Consequently, the evidential version of BIC for EGMM is then derived as
\begin{equation}
\text{EBIC}(\widetilde{\mathbf{\Theta}}_C, C)  = \log \mathcal{L}_O (\mathbf{\widetilde{\Theta}}_C) - \frac{1}{2}\widetilde{v}_{C}\log(N),
\end{equation}
where $\widetilde{v}_{C} = M-1+CD+D(D+1)/2$, including $M-1$ independent parameters in the mixing probabilities $\{\widetilde{\pi}_j\}_{j=1}^{M}$, $CD$ independent parameters in the mean vectors $\{\mu_k\}_{k=1}^{C}$, and $D(D+1)/2$ independent parameters in the common covariance matrix $\mathbf{\Sigma}$.
This index has to be maximized to determine the optimal number of clusters.

\section{Experiments} \label{sec4}
This section consists of three parts. In Section \ref{sec4.1}, some numerical examples are used to illustrate the behavior of the EGMM algorithm\footnote{The source code can be downloaded from https://github.com/jlm-138/EGMM.}. In Section \ref{sec4.2}, we compare the performance of our proposal with those of related clustering algorithms based on several real datasets. In Section \ref{sec4.3}, the interest of the proposed EGMM is illustrated by an application to multi-modal brain image segmentation.

\subsection{Illustrative examples} \label{sec4.1}
In this section, we consider three numerical examples to illustrate the interest of the proposed EGMM algorithm for deriving evidential partition that better characterizes cluster-membership uncertainty.

\subsubsection{Diamond dataset}
\begin{figure}[!ht]
\centering
  \includegraphics[scale=0.6]{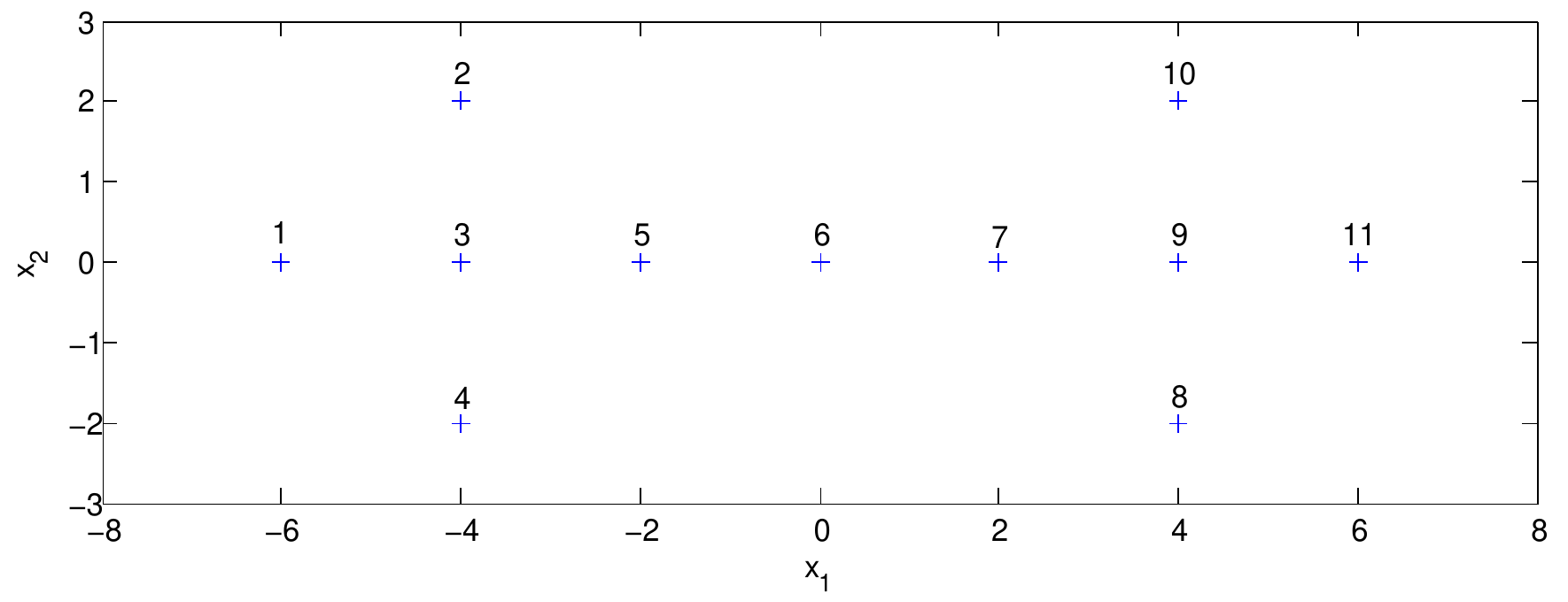}
  \caption{Diamond dataset}\label{fig_Diamond}
\end{figure}

\begin{figure}[!ht]
\centering
\subfigure[Cluster probability of each object]{
\includegraphics[scale=0.6]{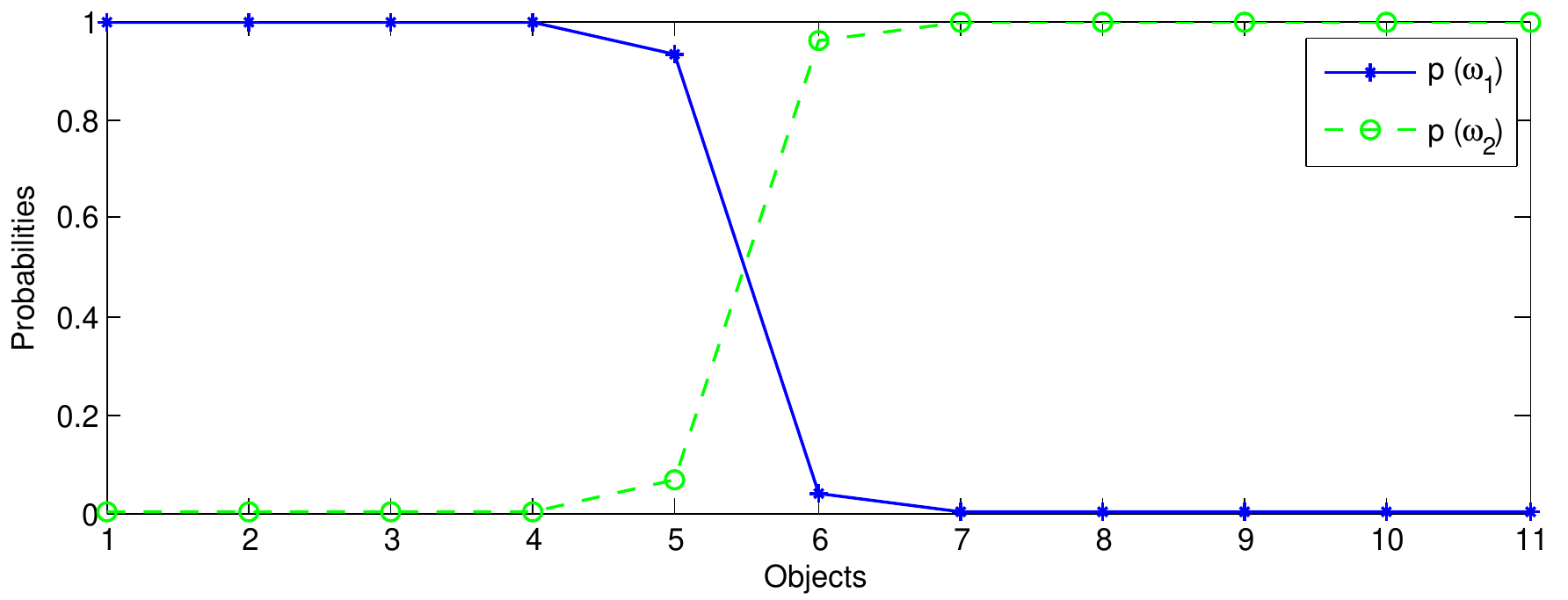}}
\subfigure[Hard partition result and the cluster centers]{
\includegraphics[scale=0.6]{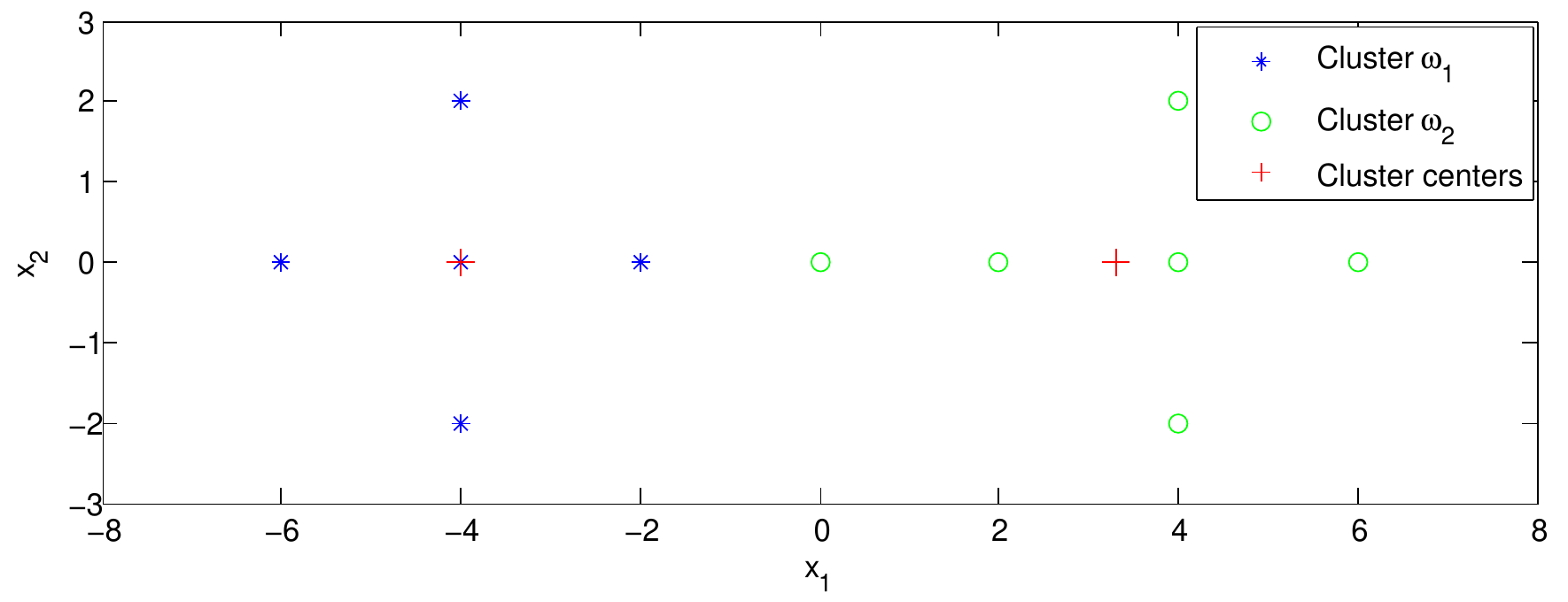}}
\caption{Diamond dataset: clustering results by GMM}
\label{fig_GMM_Diamond}
\end{figure}

\begin{figure}[!ht]
\centering
\subfigure[Cluster evidential membership of each object]{
\includegraphics[scale=0.6]{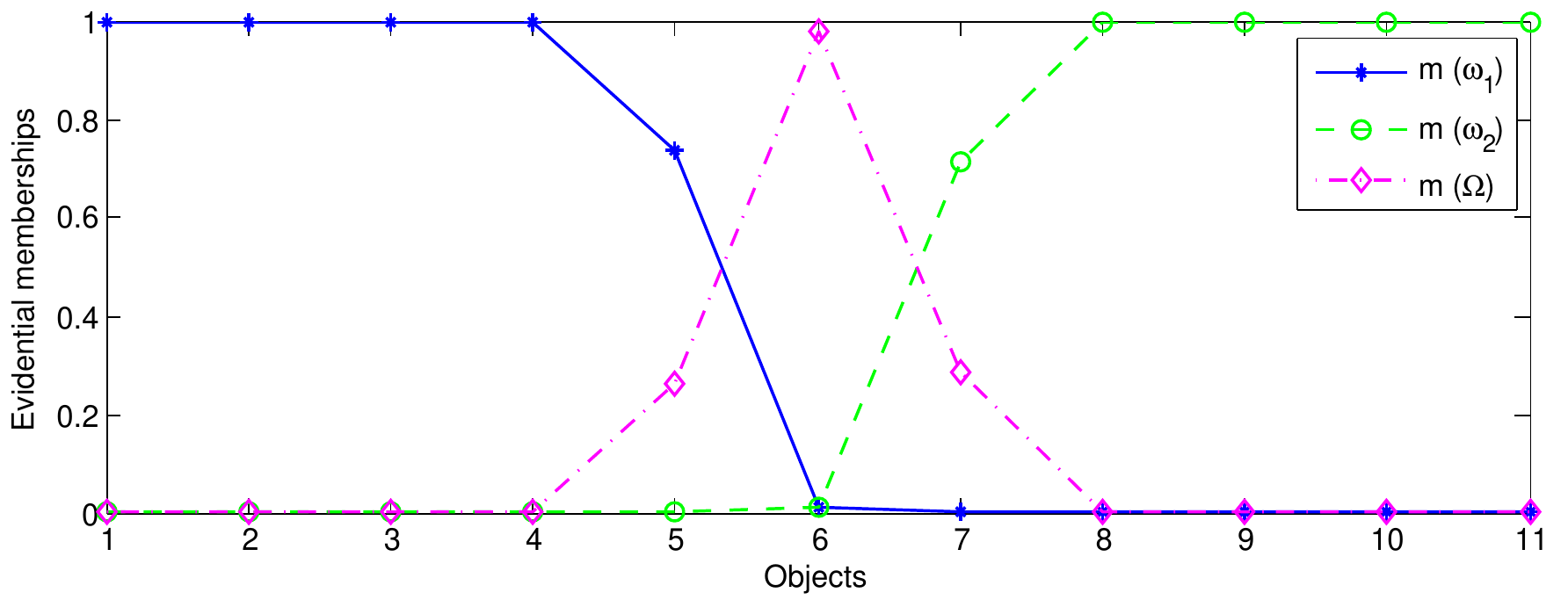}}
\subfigure[Hard evidential partition result and the cluster centers]{
\includegraphics[scale=0.6]{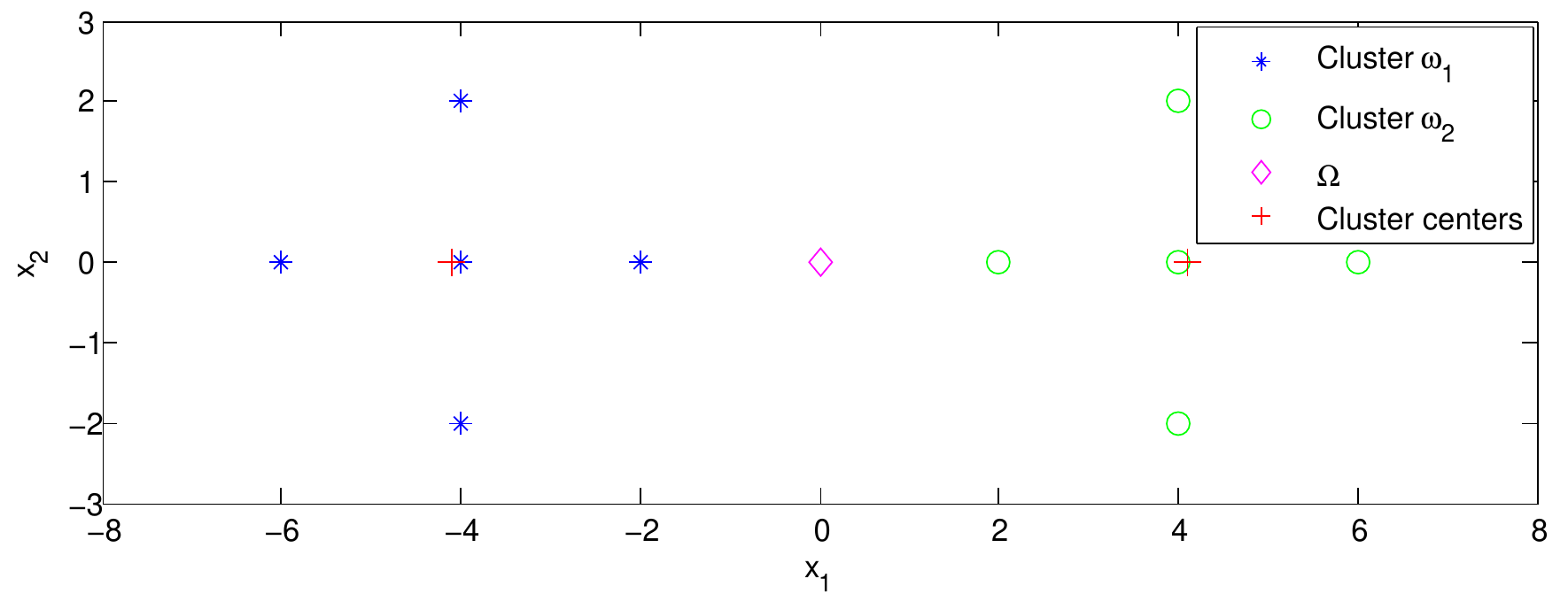}}
\caption{Diamond dataset: clustering results by EGMM}
\label{fig_EGMM_Diamond}
\end{figure}

In the first example, we consider the famous Diamond dataset to illustrate the behavior of EGMM compared with the general GMM \cite{Aggarwal14}. This dataset is composed of 11 objets, as shown in Fig. \ref{fig_Diamond}. We first calculated the cluster validity indices by running the EGMM algorithm under different numbers of clusters. Table \ref{tab_BIC_Diamond} shows the EBIC indices with the desired number of clusters ranging from 2 to 6. It can be seen that the maximum is obtained for $C = 2$ clusters, which is consistent with our intuitive understanding for the partition of this dataset. Figs. \ref{fig_GMM_Diamond} and \ref{fig_EGMM_Diamond} show the clustering results (with $C = 2$) by GMM and EGMM, respectively. For the GMM, object 6, which lies at the cluster boundary, is assigned a high probability to cluster $\omega_2$. But for our proposed EGMM, object 6 is assigned a high evidential membership to $\Omega$, which reveals that this point is ambiguous: it could be assigned either to $\omega_1$ or $\omega_2$. In addition, the EGMM can find the approximate locations for both of the two cluster centers, whereas the GMM gives a biased estimation for the center location of cluster $\omega_2$. This example demonstrates that the proposed EGMM is more powerful to detect those ambiguous objects, and thus can reveal the underlying structure of the considered data in a more comprehensive way.

\begin{table}[!ht]
\caption{Diamond dataset: EBIC indices for different numbers of clusters}
\begin{center}
\begin{tabular}{p{30mm}|p{12mm}p{12mm}p{12mm}p{12mm}p{12mm}} 
\hline
\textbf{Cluster number}    &\textbf{2}              &3          &4          &5          &6\\
\hline
\textbf{EBIC index}             &\textbf{-55.9}        &-63.1     &-70.6     &-91.5     &-132.3\\
\hline
\end{tabular}
\end{center} \label{tab_BIC_Diamond}
\end{table}

\subsubsection{Two-class dataset}
\begin{figure*}[!ht]
\centering
\subfigure[The dataset]{
\includegraphics[scale=0.58]{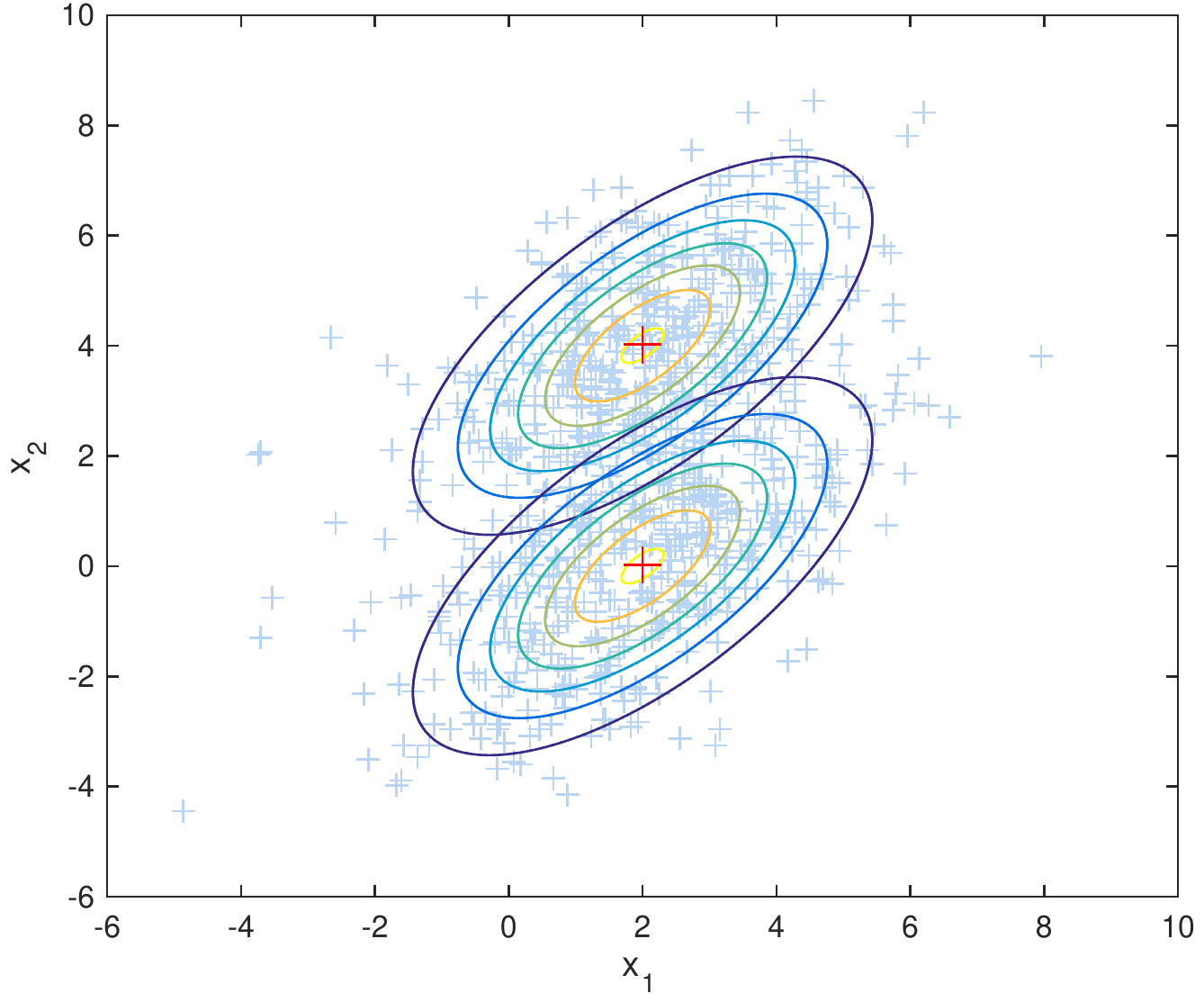}}
\subfigure[Hard evidential partition result and the cluster centers by ECM]{
\includegraphics[scale=0.58]{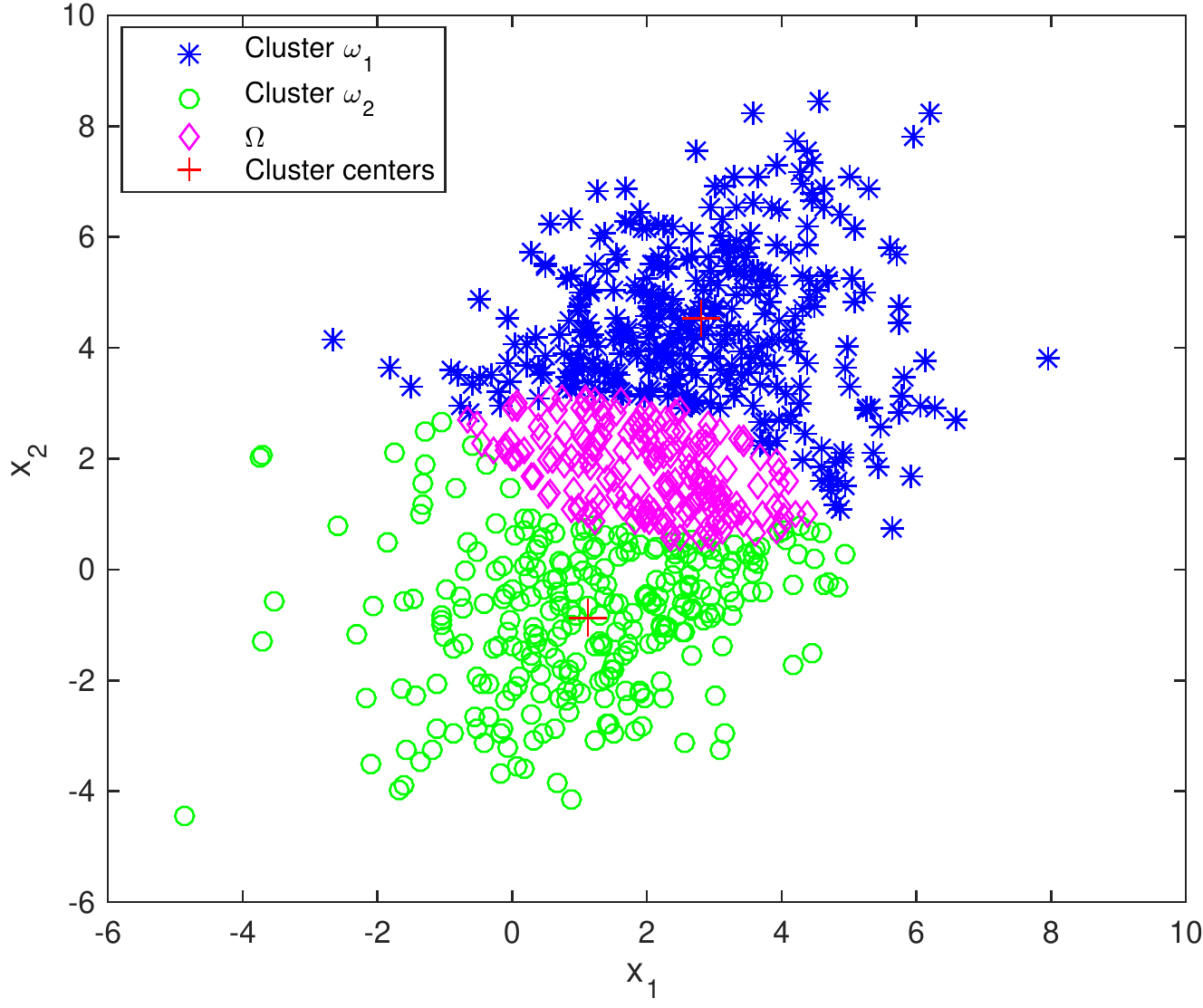}}
\subfigure[Hard evidential partition result and the cluster centers by bootGMM]{
\includegraphics[scale=0.58]{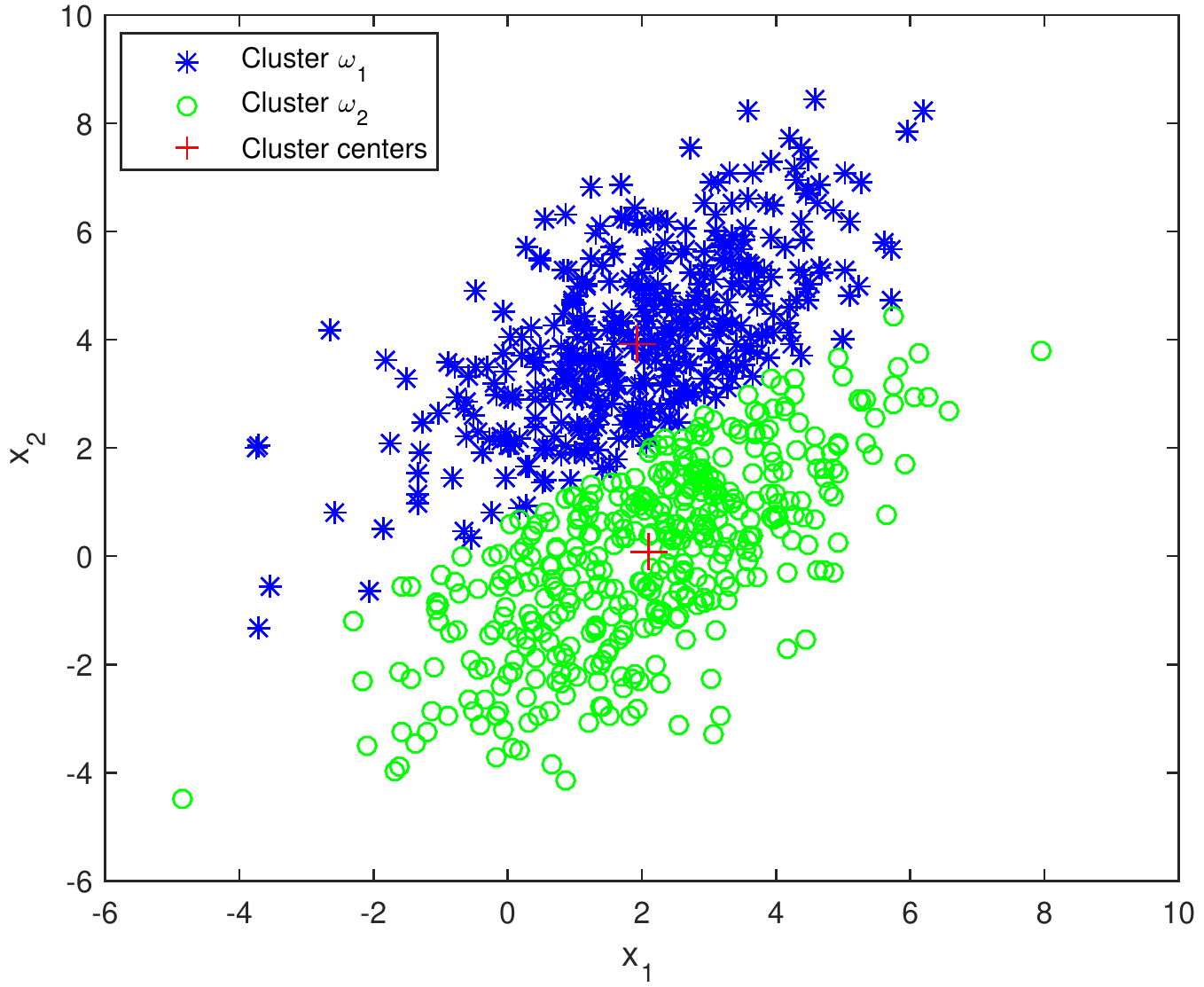}}
\subfigure[Hard evidential partition result and the cluster centers by EGMM]{
\includegraphics[scale=0.58]{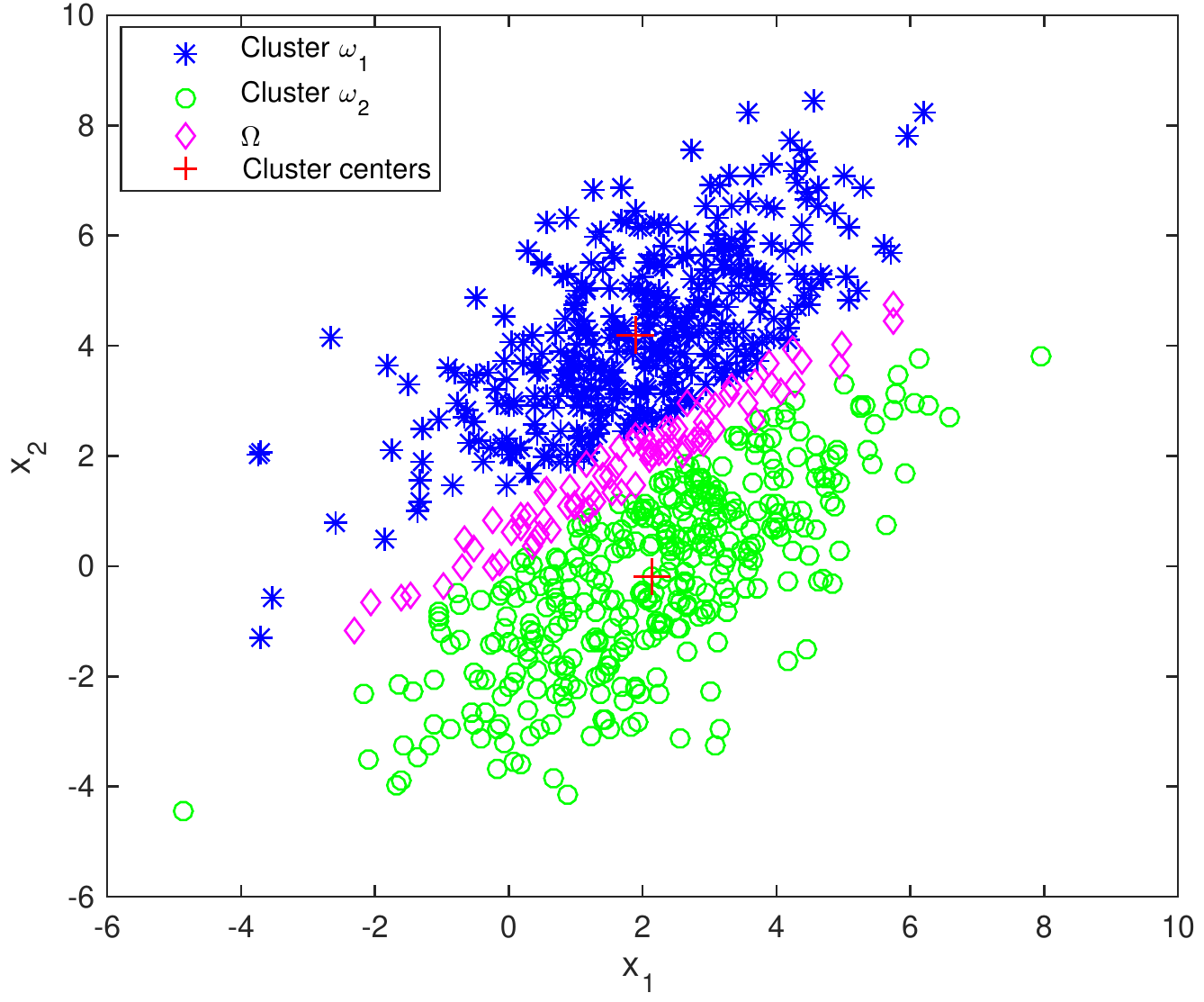}}
\caption{Two-class dataset: clustering results by ECM, bootGMM, and EGMM}
\label{fig_TwoGaussian}
\end{figure*}

In the second example, a dataset generated by two Gaussian distributions is considered to demonstrate the superiority of the proposed EGMM compared with the prototype-based ECM \cite{Masson08} and the model-based bootGMM \cite{Denoeux20}, which are two representative evidential clustering algorithms developed in the belief function framework. This dataset is composed of two classes of 400 points, generated from Gaussian distributions with the same covariance matrix $[3,2; 2,3]$ and different mean vectors, $[2,4]$ and $[2,0]$, respectively. The dataset and the contours of the distributions are represented in Fig. \ref{fig_TwoGaussian} (a). We first calculated the cluster validity indices by running the EGMM algorithm under different numbers of clusters. Table \ref{tab_BIC_Two-class} shows the EBIC indices with the desired number of clusters ranging from 2 to 6. It indicates that the number of clusters should be chosen as $C=2$, which is consistent with the real class distributions. Figs. \ref{fig_TwoGaussian} (b)-(d) show the clustering results (with $C = 2$) by ECM, bootGMM and EGMM, respectively. It can be seen that, the ECM fails to recover the underlying structure of the dataset, which is because the Euclidan distance-based similarity measure can only discover hyperspherical clusters. The proposed EGMM accurately recovers the two underlying hyperellipsoid clusters thanks to the adaptive similarity measure derived via MLE. This example demonstrates that the proposed EGMM is more powerful to distinguish hyperellipsoid clusters with arbitrary orientation and shape than ECM. As for the bootGMM, it successfully recovers the two underlying hyperellipsoid clusters by fitting the model based on mixtures of Gaussian distributions with arbitrary geometries. However, it fails to detect those ambiguous objects lying at the cluster boundary via the hard evidential partition, as quite small evidential membership is assign to $\Omega$ for these objects. By comparison, the proposed EGMM can automatically detect these ambiguous objects thanks to the mixture models constructed over the powerset of the desired clusters.

\begin{table}[!ht]
\caption{Two-class dataset: EBIC indices for different numbers of clusters}
\begin{center}
\begin{tabular}{p{30mm}|p{12mm}p{12mm}p{12mm}p{12mm}p{12mm}} 
\hline
\textbf{Cluster number}    &\textbf{2}              &3          &4          &5          &6\\
\hline
\textbf{EBIC index}            &\textbf{-3395.6}        &-3413.1    &-3425.6    &-3443.7    &-3466.8\\
\hline
\end{tabular}
\end{center} \label{tab_BIC_Two-class}
\end{table}

\subsubsection{Four-class dataset}
\begin{figure*}[!ht]
\centering
\subfigure[The dataset]{
\includegraphics[scale=0.58]{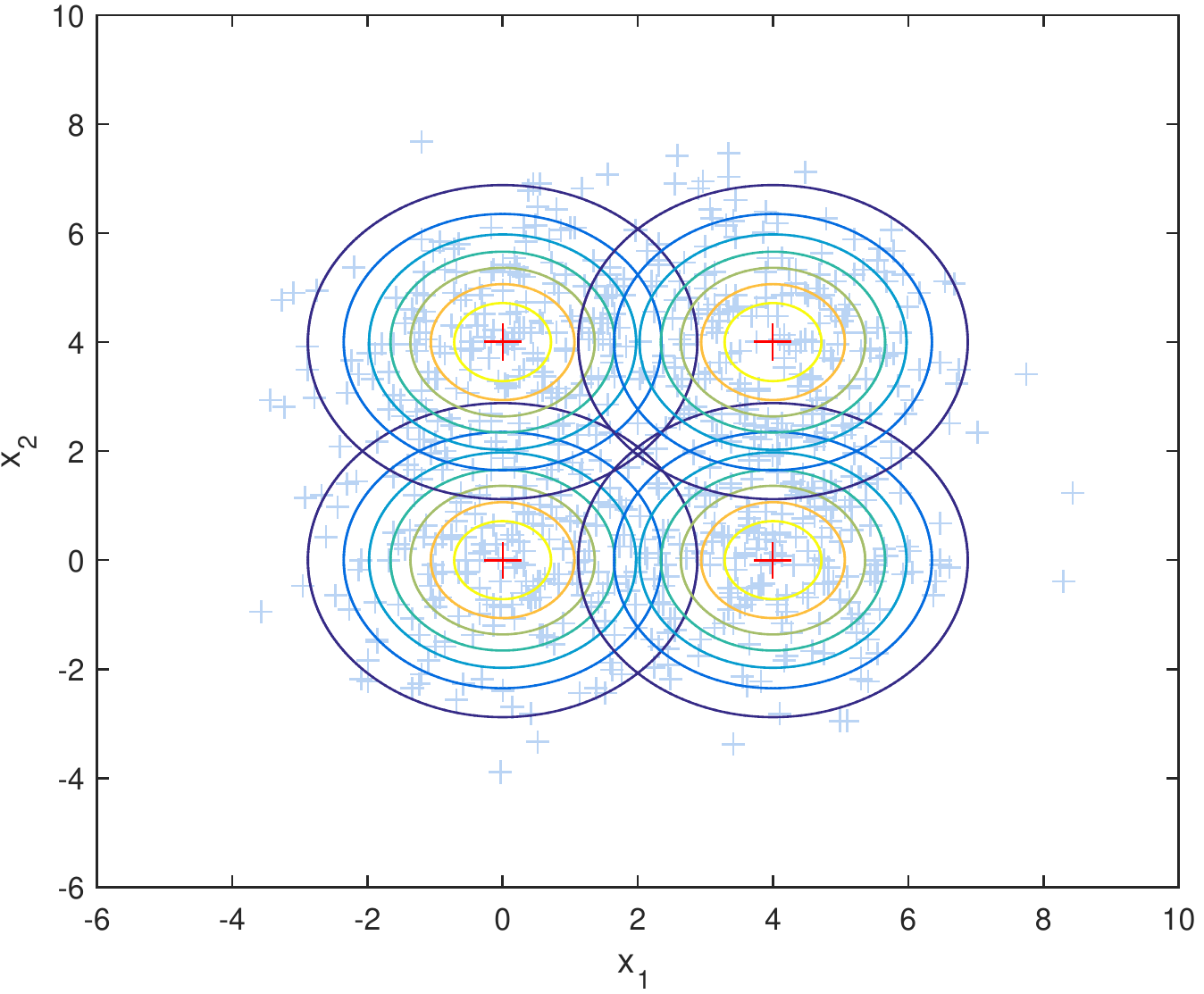}}
\subfigure[Hard evidential partition result and the cluster centers]{
\includegraphics[scale=0.58]{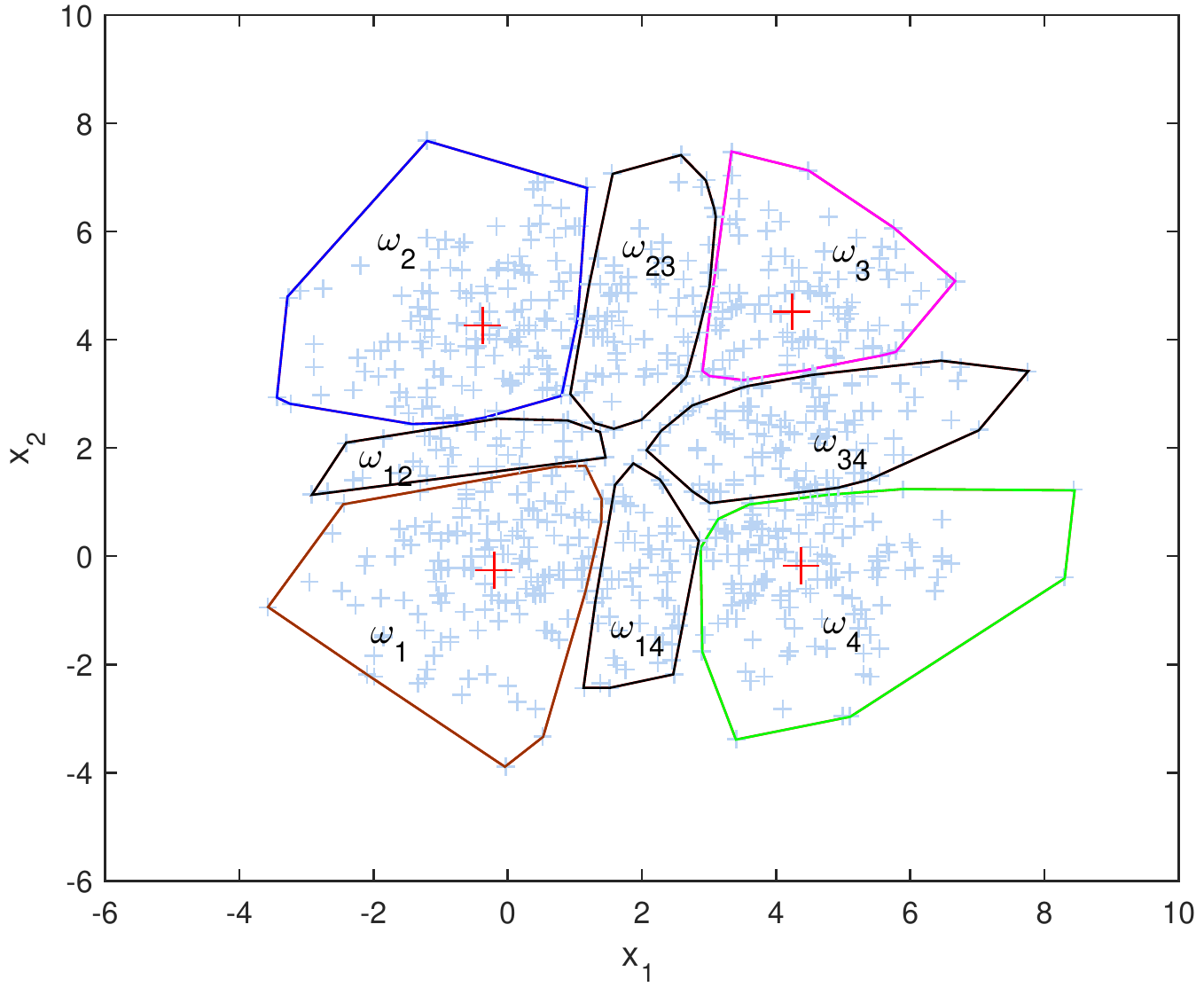}}
\subfigure[Lower approximations of the four clusters]{
\includegraphics[scale=0.58]{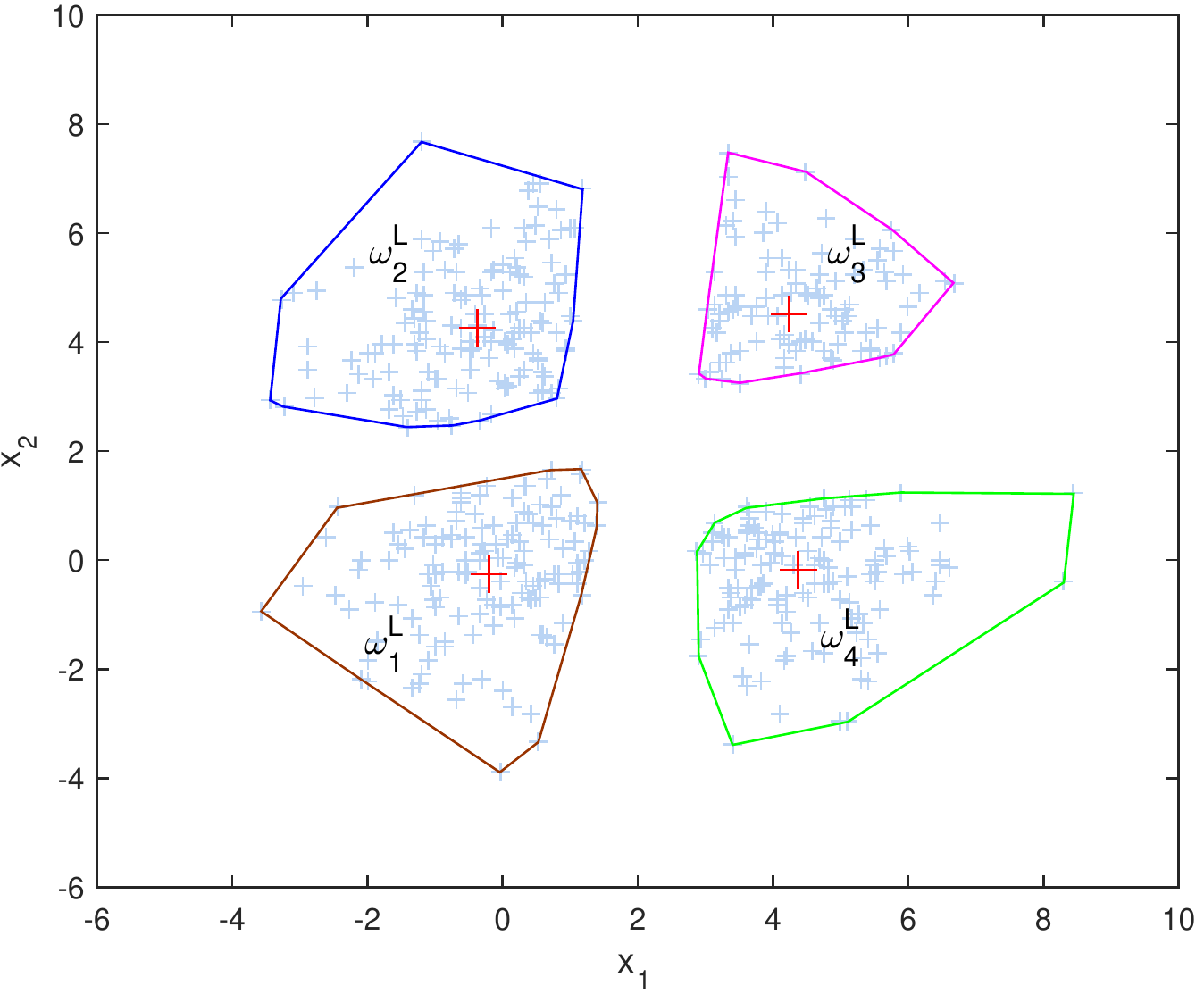}}
\subfigure[Upper approximations of the four clusters]{
\includegraphics[scale=0.58]{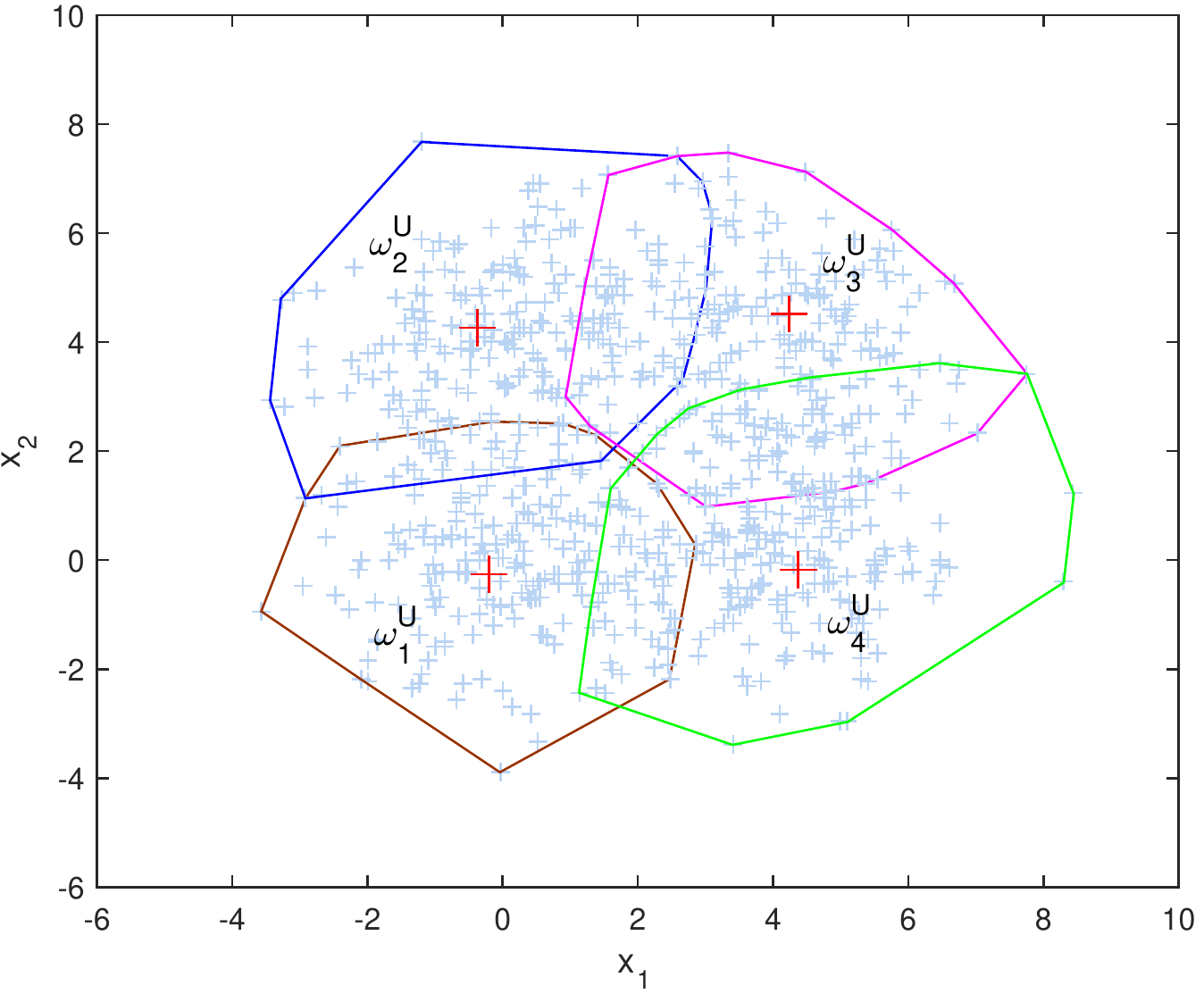}}
\caption{Four-class dataset: interpretation of the evidential partition generated by EGMM}
\label{fig_FourGaussian}
\end{figure*}

In the third example, a more complex dataset is considered to illustrate the interest of evidential partition obtained by the proposed EGMM. This dataset is composed of four classes of 200 points, generated from Gaussian distributions with the same covariance matrix $2\textbf{I}$ and different mean vectors, $[0,0]$, $[0,4]$, $[4,4]$, and $[4,0]$, respectively. The dataset and the contours of the distributions are represented in Fig. \ref{fig_FourGaussian} (a). We first calculated the cluster validity indices by running the EGMM algorithm under different numbers of clusters. Table \ref{tab_BIC_Four-class} shows the EBIC indices with the desired number of clusters ranging from 2 to 6. Noting that the maximum is obtained for $C = 4$ clusters, the underlying structure of the dataset is correctly discovered. Fig. \ref{fig_FourGaussian} (b) shows the hard evidential partition result (represented by convex hull) and the cluster centers (marked by red cross) with $C = 4$. It can be seen that the four clusters are accurately recovered, and those points that lie at the cluster boundaries are assigned to the ambiguous sets of clusters. Apart from the hard evidential partition, it is also possible to characterize each cluster $\omega_k$ by two sets: the set of objects which can be classified as $\omega_k$ without any ambiguity and the set of objects which could possibly be assigned to $\omega_k$ \cite{Masson08}. These two sets $\omega_k^L$ and $\omega_k^U$, referred to as the lower and upper approximations of $\omega_k$, are defined as $\omega_k^L = X({\omega_k})$ and $\omega_k^U = \bigcup \limits_ {j \setminus \omega_k \in A_j} X(A_j)$, with $X(A_j)$ denoting the set of objects for which the mass assigned to $A_j$ is highest. Figs. \ref{fig_FourGaussian} (c) and (d) show the lower and upper approximations of each cluster, which provide a pessimistic and an optimistic clustering results, respectively. This example demonstrates that the evidential partition generated by the proposed EGMM is quite intuitive and easier to interpret than the numerical probabilities obtained by the GMM, and can provide much richer partition information than the classical hard partition.

\begin{table}[!ht]
\caption{Four-class dataset: EBIC indices for different numbers of clusters}
\begin{center}
\begin{tabular}{p{30mm}|p{12mm}p{12mm}p{12mm}p{12mm}p{12mm}} 
\hline
\textbf{Cluster number}    &2              &3          &\textbf{4}          &5          &6\\
\hline
\textbf{EBIC index}            &-3648.5        &-3666.5    &\textbf{-3635.2}    &-3655.1    &-3679.9\\
\hline
\end{tabular}
\end{center} \label{tab_BIC_Four-class}
\end{table}

\subsection{Real data test} \label{sec4.2}
\subsubsection{Experimental setup} \label{sec4.2.1}
In this section, we aim to evaluate the performance of the proposed EGMM based on eight representative benchmark datasets from the UCI Machine Learning Repository \cite{Dua2017}, whose characteristics are summarized in Table \ref{tab_Real}. It can be seen that the selected datasets varies greatly in both size and dimensionality. The clustering results obtained with the proposed EGMM were compared with the following representative clustering algorithms:
\begin{itemize}
  \item HCM \cite{Jain10}: hard $c$-means (function \texttt{kmeans} in the MATLAB \texttt{Statistics} toolbox).
  \item FCM \cite{Bezdek81}: fuzzy $c$-means (function \texttt{fcm} in the MATLAB \texttt{Fuzzy Logic} toolbox).
  \item ECM \cite{Masson08}: evidential $c$-means (function \texttt{ECM} in the MATLAB package \texttt{Evidential Clustering}\footnote{Available at https://www.hds.utc.fr/\~{}tdenoeux/dokuwiki/en/software}).
  \item BPEC \cite{Su19}: belief-peaks evidential clustering (function \texttt{bpec} in the R package \texttt{evclust}\footnote{Available at https://cran.r-project.org/web/packages/evclust}).
  \item GMM \cite{Aggarwal14}: general Gaussian mixture model without constraints on covariance (function \texttt{fitgmdist} in the MATLAB \texttt{Statistics} toolbox).
  \item CGMM \cite{Aggarwal14}: constrained Gaussian mixture model with constant covariance across clusters (function \texttt{fitgmdist} with `SharedCovariance' = true).
  \item bootGMM \cite{Denoeux20}: calibrated model-based evidential clustering by bootstrapping the most fitted GMM (function \texttt{bootclus} in the R package \texttt{evclust}).
\end{itemize}

\begin{table}[!ht]
\caption{Characteristics of the real datasets used in the experiment}
\begin{center}
\begin{tabular}{p{25mm}p{17mm}p{17mm}p{15mm}} 
\hline
Datasets &\# Instances &\# Features &\# Clusters\\
\hline
Ecoli            &336    &5    &8   \\
Iris             &150    &4    &3   \\
Knowledge        &403    &5    &4   \\
Magic            &19020  &10   &2   \\
Newthyroid       &215    &5    &3   \\
Seeds            &210    &7    &3   \\
Vehicle          &846    &18   &4   \\
Wine             &178    &13   &3   \\
\hline
\end{tabular}
\end{center} \label{tab_Real}
\end{table}

To perform a fair evaluation of the clustering results, hard partitions were adopted for all the considered algorithms. For the four evidential clustering algorithms, hard partitions were obtained by selecting the cluster with maximum pignistic probability for each object. The following three common external criteria were used for evaluation \cite{Manning08}:
\begin{itemize}
  \item Purity: Purity is a simple and transparent evaluation measure. To compute purity, each cluster is assigned to the class which is most frequent in the cluster, and then the accuracy of this assignment is measured by counting the number of correctly assigned objects and dividing by $N$. Formally,
\begin{equation}
\text{Purity}(\Omega, \mathbb{Q}) =  \frac{1}{N} \sum \limits_{k=1} \limits^{C} \max \limits_{j} |\omega_k \cap q_j|,
\end{equation}
where $\Omega  = \{\omega_{1}, \ldots, \omega_{C}\}$ is the set of partitioned clusters and $\mathbb{Q}  = \{q_{1}, \ldots, q_{J}\}$ is the set of actual classes.
  \item NMI (Normalized Mutual Information): NMI is an information-theoretical evaluation measure, which is defined as
\begin{equation}
\text{NMI}(\Omega, \mathbb{Q}) =  \frac{I(\Omega; \mathbb{Q})}{[H(\Omega)+H(\mathbb{Q})]/2},
\end{equation}
where $I$ and $H$ denote the operations of mutual information and entropy, respectively.
  \item ARI (Adjusted Rand Index): ARI is a pair counting based evaluation measure, which is defined as
\begin{equation}
\text{ARI}(\Omega, \mathbb{Q}) =  \frac{2(\text{TP} \cdot \text{TN} - \text{FP} \cdot \text{FN})}{(\text{TN}+\text{FP})(\text{FP}+\text{TP})+(\text{TN}+\text{FN})(\text{FN}+\text{TP})},
\end{equation}
where TP, TN, FP, FN denote true positive samples, true negative samples, false positive samples and false negative samples, respectively.

\end{itemize}

\subsubsection{Comparison with unknown number of clusters} \label{sec4.2.2}
In the first group of experiments, the number of clusters $C$ was assumed to be unknown and had to be determined based on the affiliated validity indices. For determining the number of clusters, the validity indices of modified partition coefficient (MPC) \cite{Dave96} and average normalized specificity (ANS) \cite{Masson08} were used for FCM and ECM, respectively, and the classical BIC \cite{Fraley02} was used for the three model-based algorithms including GMM, CGMM and bootGMM. As for BPEC, the delta-Bel graph \cite{Su19} was used to determine the number of clusters. For all the algorithms except HCM (which requires the number of clusters to be known in advance), the number of clusters was searched between 2 and 8. All algorithms were run 10 times, and the average estimated number of clusters was calculated for each algorithm. For evaluating the clustering performance, the average NMI and ARI were calculated for each algorithm using the corresponding estimated number of clusters. Note that the purity measure was not used here because it is severely affected by the number of clusters as indicated in \cite{Manning08} (high purity is easy to achieve when the number of clusters is large).

Table \ref{tab_C} shows the number of clusters estimated by different algorithms in comparison with the real number of clusters (the second column). We can see that the proposed EGMM performed the best for determining the number of clusters (obtaining the best estimation accuracy on five of the eight datasets), while the performance of other algorithms was generally unstable. Tables \ref{tab_unNMI} and \ref{tab_unARI} show the clustering results of different algorithms with the estimated number of clusters in terms of NMI and ARI, in which the bootGMM algorithm failed on the large dataset \texttt{Magic}. The numbers in brackets in the tables represent the performance ranks of different algorithms. It can be seen that the proposed EGMM achieved the highest clustering quality as measured by NMI and ARI for most of the datasets and it ranked the first in average. These results show the superiority of the proposed EGMM both in finding the number of clusters and clustering the data.

\begin{table*}[!ht] \footnotesize
\caption{Number of clusters estimated by different algorithms}
\begin{center}
\begin{tabular}{p{20mm}p{13mm}p{13mm}p{13mm}p{13mm}p{13mm}p{13mm}p{13mm}p{11mm}} 
\hline
Dataset  &\# Clusters  &FCM  &ECM  &BPEC  &GMM  &CGMM &bootGMM &EGMM\\
\hline
Ecoli     &8  &4.0$\pm$0      &4.0$\pm$0     &3.0$\pm$0  &4.0$\pm$0   &4.5$\pm$0.71     &\textbf{5.0$\pm$0}    &4.0$\pm$0\\
Iris      &3  &4.1$\pm$1.45   &2.0$\pm$0     &2.0$\pm$0  &2.0$\pm$0   &4.5$\pm$0.85     &2.0$\pm$0    &\textbf{3.4$\pm$0.52}\\
Knowledge &4  &4.9$\pm$1.10   &2.0$\pm$0    &3.0$\pm$0   &2.5$\pm$0.53   &2.5$\pm$0.85     &3.0$\pm$0 &\textbf{3.8$\pm$0.63}\\
Magic     &2  &4.4$\pm$0.90   &5.0$\pm$0    &4.0$\pm$0   &5.0$\pm$0     &5.0$\pm$0        &-- &\textbf{3.2$\pm$0.95}\\
Newthyroid&3  &2.0$\pm$0      &6.0$\pm$0    &2.0$\pm$0   &5.8$\pm$0.42  &5.5$\pm$0.71     &\textbf{3.0$\pm$0}  &3.2$\pm$0.42\\
Seeds     &3  &4.7$\pm$1.06   &5.4$\pm$0.52 &2.0$\pm$0 &2.0$\pm$0    &3.1$\pm$0.31   &4.0$\pm$0  &\textbf{3.0$\pm$0}\\
Vehicle   &4  &4.1$\pm$1.45   &6.0$\pm$0  &\textbf{4.0$\pm$0}  &5.1$\pm$0.57     &4.7$\pm$0.95    &6.0$\pm$0 &5.9$\pm$0.32\\
Wine      &3  &4.1$\pm$1.45   &6.0$\pm$0     &2.0$\pm$0   &2.0$\pm$0     &5.1$\pm$0.88  &4.0$\pm$0  &\textbf{3.8$\pm$0.42}\\
\hline
\end{tabular}
\end{center} \label{tab_C}
\end{table*}

\begin{table*}[!ht] \footnotesize
\caption{NMI by different algorithms with unknown number of clusters}
\begin{center}
\begin{tabular}{p{21mm}p{15mm}p{15mm}p{15mm}p{15mm}p{15mm}p{15mm}p{15mm}} 
\hline
Dataset    &FCM  &ECM  &BPEC  &GMM  &CGMM &bootGMM &EGMM\\
\hline
Ecoli       &0.58$\pm$0.01(7)  &0.59$\pm$0.01(5.5)  &0.65$\pm$0(4)  &0.69$\pm$0.01(1)  &0.67$\pm$0.01(3) &0.59$\pm$0.01(5.5) &0.68$\pm$0.01(2)\\
Iris        &0.70$\pm$0.01(6)  &0.57$\pm$0.05(7) &0.72$\pm$0(5)  &0.73$\pm$0(3.5)  &0.77$\pm$0.06(2)    &0.73$\pm$0(3.5) &0.87$\pm$0.05(1)\\
Knowledge   &0.29$\pm$0.02(5)  &0.33$\pm$0(4)   &0.16$\pm$0(6)   &0.36$\pm$0.12(3)  &0.02$\pm$0.02(7)    &0.39$\pm$0(2) &0.43$\pm$0.05(1)\\
Magic       &0.01$\pm$0(5.5)   &0.01$\pm$0(5.5)   &0.06$\pm$0(4)   &0.12$\pm$0.01(2)  &0.11$\pm$0.03(3)    &-- --  &0.13$\pm$0(1)\\
Newthyroid  &0.13$\pm$0(7)     &0.34$\pm$0(5)   &0.22$\pm$0(6)   &0.46$\pm$0.01(4)  &0.50$\pm$0.08(2)    &0.68$\pm$0.01(1) &0.48$\pm$0.05(3)\\
Seeds       &0.61$\pm$0(3)     &0.58$\pm$0(6)    &0.50$\pm$0(7) &0.60$\pm$0(4)   &0.78$\pm$0.07(2)  &0.59$\pm$0.01(5)  &0.80$\pm$0(1)\\
Vehicle     &0.18$\pm$0(4)     &0.12$\pm$0(7)    &0.13$\pm$0(6)  &0.19$\pm$0.03(3)    &0.15$\pm$0.04(5)  &0.35$\pm$0.01(1) &0.24$\pm$0(2)\\
Wine        &0.36$\pm$0(5)     &0.35$\pm$0.01(6) &0.16$\pm$0(7)  &0.58$\pm$0.07(4) &0.81$\pm$0.03(3) &0.95$\pm$0(1)  &0.82$\pm$0.05(2)\\
\hline
Average rank   &5.32              &5.75             &5.63           &3.06             &3.38          &3.25        &1.62\\
\hline
\end{tabular}
\end{center} \label{tab_unNMI}
\end{table*}

\begin{table*}[!ht] \footnotesize
\caption{ARI by different algorithms with unknown number of clusters}
\begin{center}
\begin{tabular}{p{21mm}p{15mm}p{15mm}p{15mm}p{15mm}p{15mm}p{15mm}p{15mm}} 
\hline
Dataset    &FCM  &ECM  &BPEC  &GMM  &CGMM &bootGMM &EGMM\\
\hline
Ecoli       &0.51$\pm$0.04(6.5)  &0.51$\pm$0.04(6.5) &0.70$\pm$0(4)  &0.73$\pm$0.01(1)  &0.71$\pm$0.01(3)    &0.61$\pm$0.01(5) &0.72$\pm$0.01(2)\\
Iris        &0.62$\pm$0.02(3)  &0.54$\pm$0(7)    &0.56$\pm$0(6)  &0.57$\pm$0(4.5)   &0.72$\pm$0.11(2)    &0.57$\pm$0(4.5) &0.85$\pm$0.10(1)\\
Knowledge   &0.22$\pm$0.02(5)  &0.28$\pm$0(3)   &0.12$\pm$0(6)   &0.26$\pm$0.11(4)  &0.01$\pm$0(7)       &0.29$\pm$0(2) &0.31$\pm$0.04(1)\\
Magic       &0.02$\pm$0(5.5)   &0.02$\pm$0.01(5.5)  &0.03$\pm$0(4)   &0.11$\pm$0.01(2) &0.10$\pm$0.06(3)   &-- --  &0.14$\pm$0.03(1)\\
Newthyroid  &0.05$\pm$0.01(7)  &0.21$\pm$0(6)   &0.22$\pm$0(5)   &0.44$\pm$0.05(4)  &0.51$\pm$0.12(3)    &0.71$\pm$0.01(1)  &0.54$\pm$0.06(2)\\
Seeds       &0.52$\pm$0(4.5)   &0.52$\pm$0(4.5)   &0.45$\pm$0(7)  &0.51$\pm$0(6)    &0.81$\pm$0.13(2)  &0.53$\pm$0.03(3)  &0.85$\pm$0(1)\\
Vehicle     &0.12$\pm$0(5)     &0.14$\pm$0(2)    &0.09$\pm$0(7)  &0.13$\pm$0.03(3.5)    &0.10$\pm$0.03(6)  &0.21$\pm$0.01(1) &0.13$\pm$0(3.5)\\
Wine        &0.27$\pm$0(5)     &0.24$\pm$0(6)    &0.15$\pm$0(7)  &0.44$\pm$0.06(4)    &0.80$\pm$0.05(3)  &0.97$\pm$0(1)  &0.83$\pm$0.06(2)\\
\hline
Average rank   &5.19              &5.06             &5.75           &3.63             &3.63          &3.06        &1.68\\
\hline
\end{tabular}
\end{center} \label{tab_unARI}
\end{table*}

\begin{figure*}[!ht]
\centering
\subfigure[NMI criterion]{
\includegraphics[scale=1]{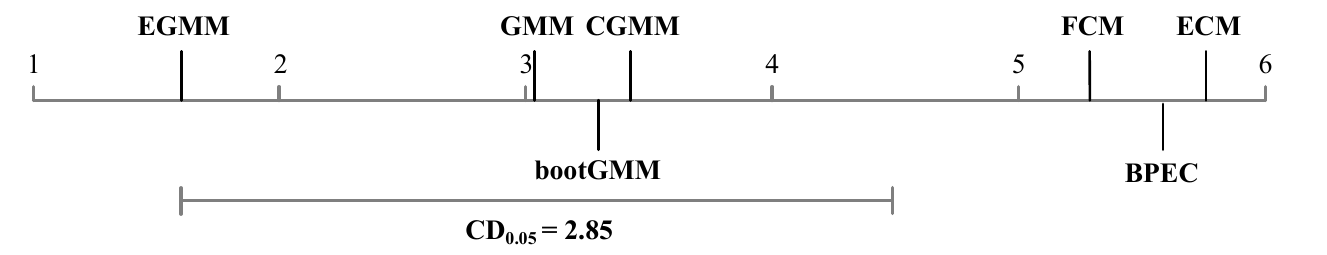}}
\subfigure[ARI criterion]{
\includegraphics[scale=1]{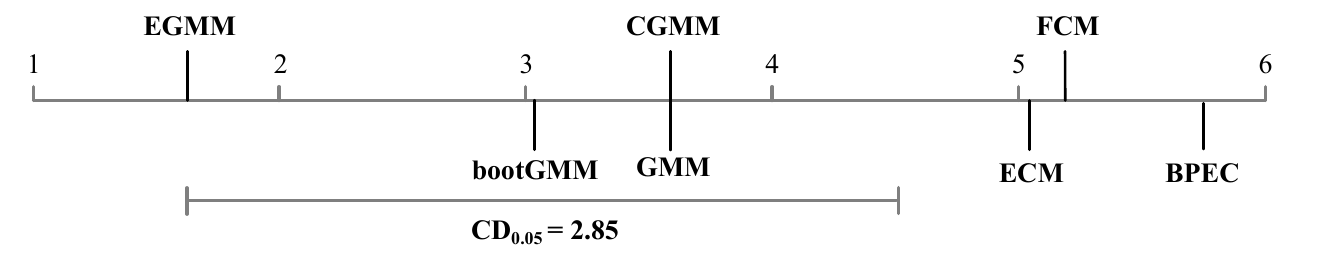}}
\caption{Bonferroni-Dun test results of EGMM versus other ones with unknown number of clusters ($\alpha=0.05$)}
\label{fig_unBD}
\end{figure*}

To analyze the results statistically, we used the Friedman test and Bonferroni-Dun test \cite{Demsar06} to compare the average ranks of different algorithms under each criterion. We first considered the Friedman test to check whether there is a significant difference among algorithms on the whole. As shown in Table \ref{tab_unFriedman}, the Friedman statistic is larger than the critical value under significance level $\alpha$=0.05 for the two criteria NMI and ARI, indicating that there is a significant difference among the results of different algorithms. Then, we applied the \emph{post hoc} Bonferroni-Dun test to compare the algorithm with highest rank (the proposed EGMM) with the other ones based on the two criteria NMI and ARI. Fig. \ref{fig_unBD} graphically shows the test results of the average ranks under significance level $\alpha=0.05$. The critical difference (CD) value is marked as horizontal line, and any one with the rank outside these areas is significantly different from the proposed EGMM. It can be seen that the proposed EGMM performed significantly better than FCM, ECM and BPEC on both of the two criteria NMI and ARI.

\begin{table*}[!ht]
\caption{Friedman test results of different algorithms with unknown number of clusters ($\alpha=0.05$)}
\begin{center}
\begin{tabular}{p{21mm}p{25mm}p{21mm}} 
\hline
Criterion    &Friedman statistic  &Critical value  \\
\hline
NMI          &25.78               &12.59  \\
ARI          &20.81               &12.59  \\
\hline
\end{tabular}
\end{center} \label{tab_unFriedman}
\end{table*}

\subsubsection{Comparison with known number of clusters} \label{sec4.2.3}
In the second group of experiments, the number of clusters was assumed to be known in advance. All algorithms were run 10 times using the real number of clusters, and the average purity, NMI, and ARI were calculated for each algorithm as shown in Tables \ref{tab_purity}-\ref{tab_ARI}, in which the bootGMM algorithm failed on the large dataset \texttt{Magic}. It can be seen that the proposed EGMM achieves the highest clustering quality as measured by purity, NMI and ARI for most of the datasets and it also ranks the first in average.

\begin{table*}[!ht] \footnotesize
\caption{Purity by different algorithms with known number of clusters}
\begin{center}
\begin{tabular}{p{15mm}p{14mm}p{14mm}p{14mm}p{14mm}p{14mm}p{14mm}p{14mm}p{14mm}} 
\hline
Dataset    &HCM  &FCM  &ECM  &BPEC  &GMM  &CGMM &bootGMM &EGMM\\
\hline
Ecoli       &0.81$\pm$0.02(2.5)  &0.79$\pm$0.01(5) &0.79$\pm$0.01(5)  &0.76$\pm$0(8)  &0.78$\pm$0.01(7)  &0.81$\pm$0.01(2.5)  &0.79$\pm$0.01(5)  &0.82$\pm$0.01(1)\\
Iris        &0.87$\pm$0.07(6.5)  &0.89$\pm$0(4)    &0.89$\pm$0(4)    &0.69$\pm$0(8)  &0.89$\pm$0.15(4)    &0.87$\pm$0.14(6.5)    &0.97$\pm$0.01(1)  &0.93$\pm$0.09(2)\\
Knowledge   &0.57$\pm$0.05(3)   &0.50$\pm$0.01(5) &0.51$\pm$0.01(4) &0.49$\pm$0(6.5)  &0.62$\pm$0.02(2) &0.38$\pm$0.10(8)    &0.49$\pm$0.01(6.5)  &0.63$\pm$0.04(1)\\
Magic       &0.65$\pm$0(5.5)     &0.65$\pm$0(5.5)   &0.65$\pm$0(5.5)   &0.65$\pm$0(5.5)   &0.66$\pm$0(3)   &0.72$\pm$0.01(2)  &-- --           &0.75$\pm$0(1)\\
Newthyroid  &0.81$\pm$0.04(5)   &0.79$\pm$0.01(6)  &0.70$\pm$0(8)   &0.77$\pm$0(7)  &0.87$\pm$0(3)  &0.86$\pm$0.02(4)   &0.91$\pm$0.01(1)  &0.88$\pm$0.03(2)\\
Seeds       &0.89$\pm$0(5)      &0.90$\pm$0(3)     &0.89$\pm$0(5)    &0.65$\pm$0(8)   &0.84$\pm$0.09(7)  &0.92$\pm$0.09(2)   &0.89$\pm$0.01(5)  &0.95$\pm$0(1)\\
Vehicle     &0.44$\pm$0.01(3.5)  &0.45$\pm$0(2)   &0.40$\pm$0(7)   &0.38$\pm$0(8)    &0.43$\pm$0.03(5)  &0.41$\pm$0.03(6)   &0.44$\pm$0.01(3.5) &0.46$\pm$0.18(1)\\
Wine        &0.69$\pm$0.01(6)   &0.69$\pm$0(6)    &0.69$\pm$0(6)    &0.56$\pm$0(8)   &0.87$\pm$0.12(2)  &0.84$\pm$0.13(4)  &0.98$\pm$0.01(1)  &0.85$\pm$0.13(3)\\
\hline
Average rank    &4.63         &4.56             &5.56           &7.38             &4.13         &4.38        &3.88        &1.50\\
\hline
\end{tabular}
\end{center} \label{tab_purity}
\end{table*}

\begin{table*}[!ht] \footnotesize
\caption{NMI by different algorithms with known number of clusters}
\begin{center}
\begin{tabular}{p{15mm}p{14mm}p{14mm}p{14mm}p{14mm}p{14mm}p{14mm}p{14mm}p{14mm}} 
\hline
Dataset    &HCM  &FCM  &ECM  &BPEC  &GMM  &CGMM &bootGMM &EGMM\\
\hline
Ecoli       &0.58$\pm$0.02(6)  &0.55$\pm$0.01(7) &0.54$\pm$0.01(8)  &0.63$\pm$0(4)   &0.69$\pm$0.01(2)  &0.68$\pm$0.01(3)  &0.61$\pm$0.01(5) &0.70$\pm$0.01(1)\\
Iris        &0.74$\pm$0(7)     &0.75$\pm$0(6)    &0.76$\pm$0(5)     &0.58$\pm$0(8)    &0.84$\pm$0.13(3)  &0.83$\pm$0.11(4)  &0.91$\pm$0.01(1)   &0.87$\pm$0.05(2)\\
Knowledge   &0.36$\pm$0.03(3)  &0.29$\pm$0.03(4.5) &0.29$\pm$0.02(4.5) &0.15$\pm$0(7)  &0.39$\pm$0.02(2)  &0.10$\pm$0.17(8)    &0.26$\pm$0.01(6)   &0.43$\pm$0.05(1)\\
Magic       &0.02$\pm$0(4)     &0.01$\pm$0(6)     &0.01$\pm$0(6)   &0.01$\pm$0(6)    &0.04$\pm$0(3)       &0.08$\pm$0.01(2)   &-- --           &0.11$\pm$0(1)\\
Newthyroid  &0.35$\pm$0.09(4.5) &0.35$\pm$0.01(4.5)  &0.16$\pm$0(8)   &0.26$\pm$0(7) &0.43$\pm$0.01(3)   &0.29$\pm$0.07(6)   &0.68$\pm$0.01(1)   &0.48$\pm$0.05(2)\\
Seeds       &0.70$\pm$0.01(3)  &0.69$\pm$0(4.5)     &0.66$\pm$0(7)    &0.43$\pm$0(8)   &0.68$\pm$0.08(6)  &0.78$\pm$0.07(2)   &0.69$\pm$0.02(4.5)  &0.80$\pm$0(1)\\
Vehicle     &0.19$\pm$0.01(3)  &0.18$\pm$0(4)     &0.13$\pm$0(7.5)    &0.13$\pm$0(7.5) &0.17$\pm$0.04(5.5)  &0.17$\pm$0.03(5.5)    &0.20$\pm$0.01(2)    &0.21$\pm$0.02(1)\\
Wine        &0.42$\pm$0.01(5.5)  &0.42$\pm$0(5.5)  &0.40$\pm$0(7)   &0.16$\pm$0(8)   &0.70$\pm$0.19(4)    &0.73$\pm$0.15(3)   &0.95$\pm$0.01(1)    &0.81$\pm$0.14(2)\\
\hline
Average rank    &4.50          &5.25             &6.63           &6.94           &3.56         &4.19        &3.56        &1.37\\
\hline
\end{tabular}
\end{center} \label{tab_NMI}
\end{table*}

\begin{table*}[!ht] \footnotesize
\caption{ARI by different algorithms with known number of clusters}
\begin{center}
\begin{tabular}{p{15mm}p{14mm}p{14mm}p{14mm}p{14mm}p{14mm}p{14mm}p{14mm}p{14mm}} 
\hline
Dataset    &HCM  &FCM  &ECM  &BPEC  &GMM  &CGMM &bootGMM &EGMM\\
\hline
Ecoli       &0.43$\pm$0.03(6)  &0.35$\pm$0.01(7.5) &0.35$\pm$0.01(7.5)  &0.69$\pm$0(4)  &0.72$\pm$0.01(1.5)  &0.71$\pm$0.01(3)  &0.64$\pm$0.01(5)   &0.72$\pm$0.01(1.5)\\
Iris        &0.70$\pm$0.10(7)  &0.73$\pm$0(5.5)    &0.73$\pm$0(5.5)    &0.51$\pm$0(8)  &0.81$\pm$0.21(3)    &0.78$\pm$0.21(4)    &0.92$\pm$0.01(1)   &0.85$\pm$0.10(2)\\
Knowledge   &0.25$\pm$0.03(3)  &0.23$\pm$0.03(4.5)  &0.23$\pm$0.03(4.5) &0.10$\pm$0(7)  &0.28$\pm$0.02(2)   &0.07$\pm$0.14(8)    &0.21$\pm$0.01(6)   &0.31$\pm$0.04(1)\\
Magic       &0.06$\pm$0(4)     &0.02$\pm$0(6.5)     &0.02$\pm$0(6.5)   &0.04$\pm$0(5)    &0.09$\pm$0(3)       &0.10$\pm$0.07(2)  &-- --           &0.14$\pm$0(1)\\
Newthyroid  &0.36$\pm$0.19(6)  &0.45$\pm$0.01(3.5)  &0.45$\pm$0.02(3.5) &0.26$\pm$0(8)  &0.42$\pm$0.05(5)   &0.27$\pm$0.07(7)   &0.71$\pm$0.01(1)  &0.54$\pm$0.06(2)\\
Seeds       &0.71$\pm$0(6)     &0.72$\pm$0(4)      &0.72$\pm$0(4)      &0.42$\pm$0(8)    &0.65$\pm$0.11(7)  &0.81$\pm$0.13(2)    &0.72$\pm$0.02(4)  &0.85$\pm$0(1)\\
Vehicle     &0.12$\pm$0(4.5)   &0.12$\pm$0(4.5)     &0.12$\pm$0(4.5)   &0.09$\pm$0(8)    &0.12$\pm$0.03(4.5)  &0.10$\pm$0.02(7)    &0.16$\pm$0.01(1)  &0.14$\pm$0.02(2)\\
Wine        &0.36$\pm$0.02(5)  &0.35$\pm$0(6.5)     &0.35$\pm$0(6.5)    &0.14$\pm$0(8)  &0.70$\pm$0.19(3)    &0.69$\pm$0.20(4)    &0.97$\pm$0.01(1)  &0.75$\pm$0.21(2)\\
\hline
Average rank    &5.19          &5.31             &5.31           &7.00           &3.63         &4.63        &3.38        &1.56\\
\hline
\end{tabular}
\end{center} \label{tab_ARI}
\end{table*}

\begin{figure*}[!ht]
\centering
\subfigure[Purity criterion]{
\includegraphics[scale=1]{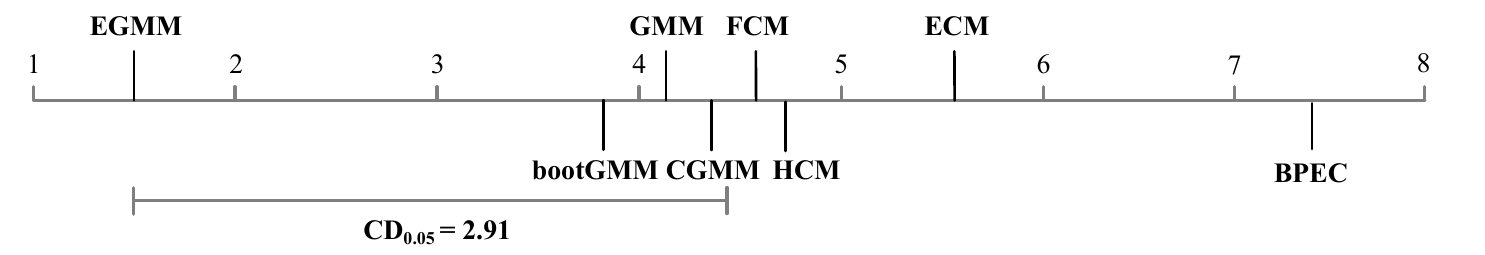}}
\subfigure[NMI criterion]{
\includegraphics[scale=1]{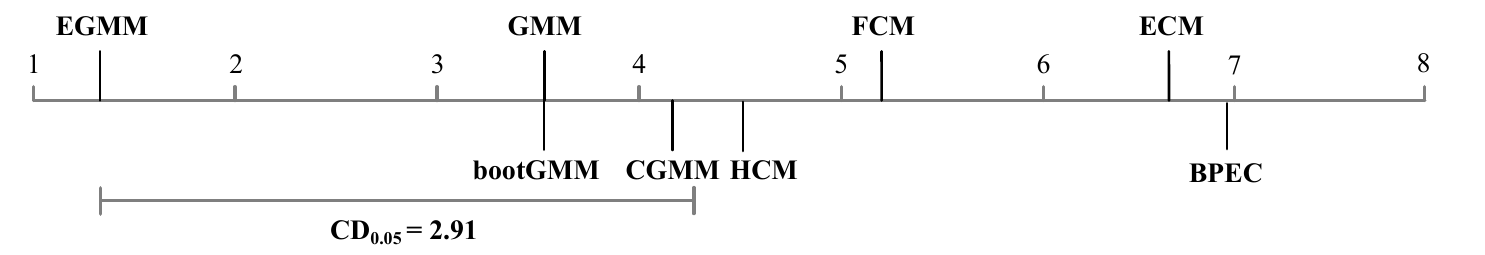}}
\subfigure[ARI criterion]{
\includegraphics[scale=1]{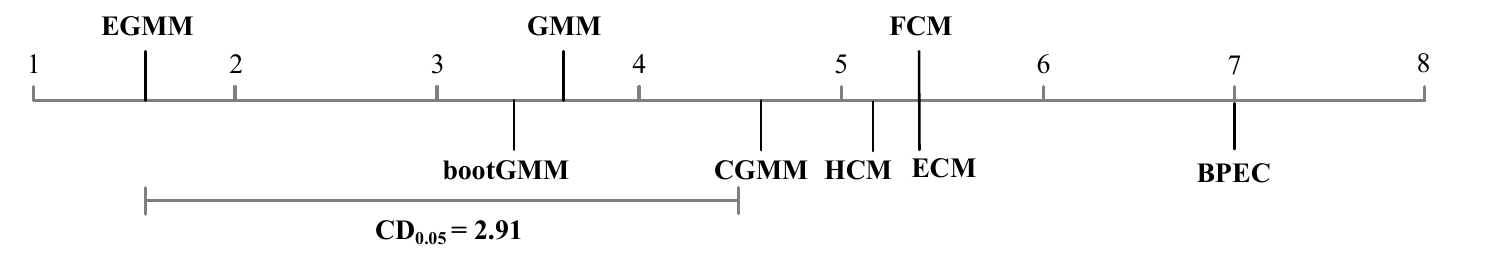}}
\caption{Bonferroni-Dun test results of EGMM versus other ones with known number of clusters ($\alpha=0.05$)}
\label{fig_BD}
\end{figure*}

We also conducted a non-parametric statistical test to compare the average ranks of different algorithms with known number of clusters. Table \ref{tab_Friedman} shows the Friedman test results of different algorithms with known number of clusters under significance level $\alpha$=0.05. As the Friedman statistics are larger than the critical value, these algorithms perform significantly different on the three criteria purity, NMI and ARI. Furthermore, Fig. \ref{fig_BD} shows the Bonferroni-Dun test results of EGMM versus other ones. It can be seen that the proposed EGMM performed significantly better than HCM, FCM, ECM and BPEC on the criteria purity and NMI, and it also performed significantly better than CGMM on the criteria ARI.

\begin{table*}[!ht]
\caption{Friedman test results of different algorithms with known number of clusters ($\alpha=0.05$)}
\begin{center}
\begin{tabular}{p{21mm}p{25mm}p{21mm}} 
\hline
Criterion    &Friedman statistic  &Critical value  \\
\hline
Purity       &24.77               &14.07  \\
NMI          &30.28               &14.07  \\
ARI          &25.07               &14.07  \\
\hline
\end{tabular}
\end{center} \label{tab_Friedman}
\end{table*}

\subsubsection{Computation time analysis} \label{sec4.2.4}
At the end of this section, we give an evaluation on the run time of the considered clustering algorithms. The computations were executed on a Microsoft Surface Book with an Intel(R) Core(TM) i5-6300U CPU @2.40 GHz and 8 GB memory. All algorithms were tested on MATLAB platform, except BPEC and bootGMM which were tested on R platform. As both of MATLAB and R are script languages, their execution efficiency is nearly at the same level. Table \ref{tab_CPU} shows the average run time of different algorithms on the eight considered datasets, in which the bootGMM algorithm failed on the large dataset \texttt{Magic}. It can be seen that the four evidential clustering algorithms (i.e., ECM, BPEC, bootGMM, EGMM) generally cost more time than the non-evidential ones, mainly because more independent parameters are needed to be estimated in the evidential partition. Among these four evidential clustering algorithms, the proposed EGMM runs fastest. In particular, it shows obvious advantage over bootGMM, which costs a great deal of time in the procedures of bootstrapping and calibration.

\begin{table*}[!ht] \small
\caption{CPU time (second) on different datasets}
\begin{center}
\begin{tabular}{p{18mm}p{14mm}p{14mm}p{14mm}p{14mm}p{14mm}p{14mm}p{14mm}p{10mm}} 
\hline
Dataset   &HCM    &FCM    &ECM    &BPEC     &GMM    &CGMM   &bootGMM      &EGMM\\
\hline
Ecoli      &0.066  &0.108  &1.486  &6.160   &0.237  &0.206  &29.35       &1.297     \\
Iris       &0.004  &0.004  &0.085  &1.970   &0.012  &0.008  &7.610       &0.029     \\
Knowledge  &0.004  &0.014  &0.080  &1.780   &0.030  &0.031  &56.66       &0.035     \\
Magic      &3.449  &4.022  &239.8  &791.6   &4.277  &3.446  &-- --       &66.91     \\
Newthyroid &0.012  &0.029  &0.410  &2.160   &0.021  &0.014  &11.42       &0.074     \\
Seeds      &0.003  &0.003  &0.078  &1.860   &0.013  &0.011  &18.16       &0.045     \\
Vehicle    &0.006  &0.022  &4.134  &55.92   &0.048  &0.034  &167.79      &2.255     \\
Wine       &0.003  &0.005  &0.116  &1.700   &0.009  &0.006  &46.77       &0.052     \\
\hline
\end{tabular}
\end{center} \label{tab_CPU}
\end{table*}

\subsection{Application to multi-modal brain image segmentation} \label{sec4.3}
\begin{figure}[!ht]
\centering
\subfigure[Brain image with T1 pulse sequence]{
\includegraphics[scale=0.6]{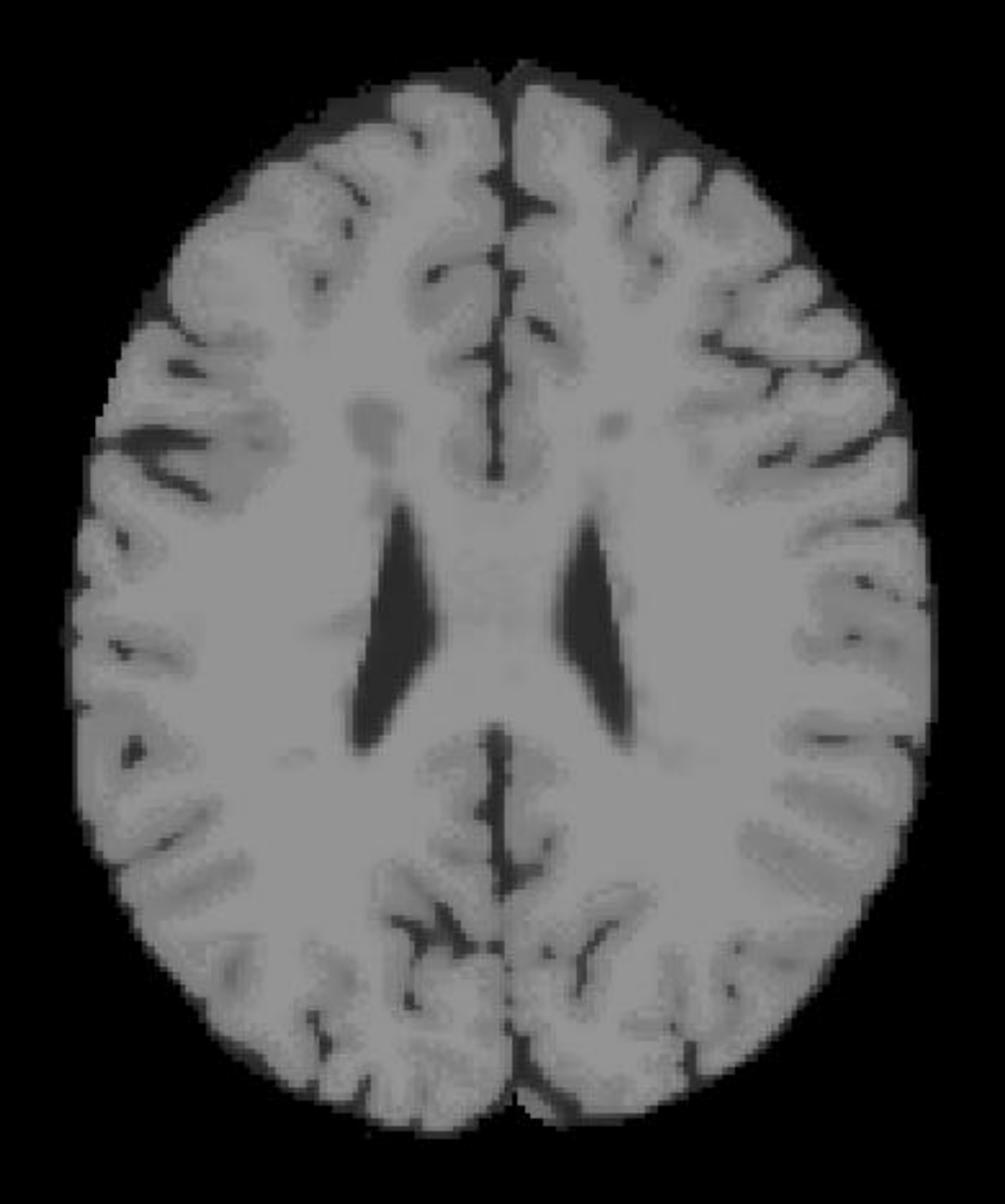}}
\subfigure[Brain image with T2 pulse sequence]{
\includegraphics[scale=0.6]{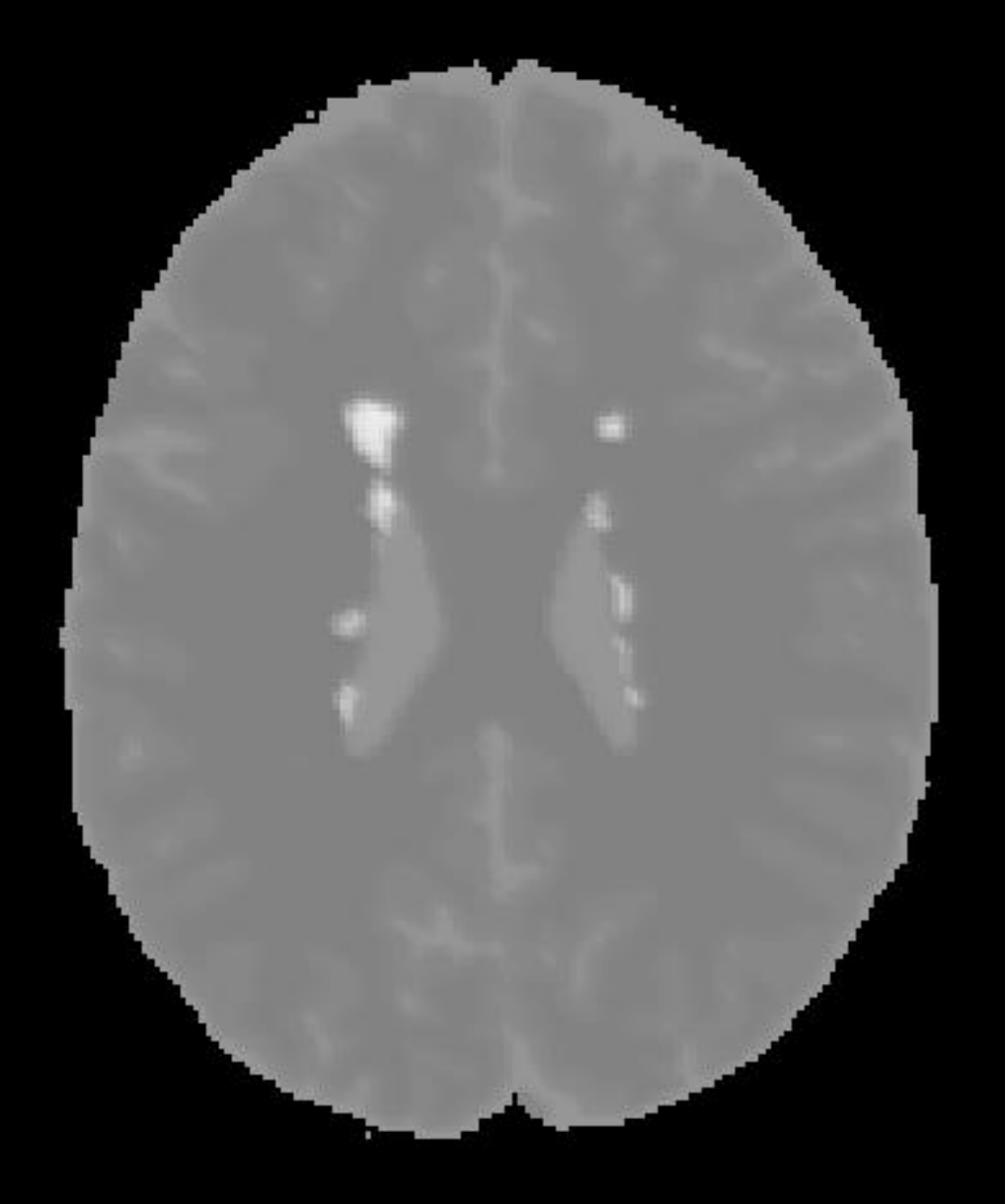}}
\caption{Multi-modal brain images with MS lesions}
\label{fig_MRI}
\end{figure}

In this section, the interest of the proposed EGMM was illustrated by an application to multi-modal brain image segmentation. The multi-modal brain images with multiple sclerosis (MS) lesions were generated using the BrainWEB tool\footnote{https://brainweb.bic.mni.mcgill.ca/brainweb/.}, in which two different pulse sequences (T1 and T2) were employed for magnetic resonance imaging (MRI). As shown in Fig. \ref{fig_MRI}(a), two main areas are easily distinguished from the T1 image: the bright area represents brain tissues, and the dark grey area represents ventricles and cerebrospinal fluid. As the T1 pulse sequence is designed for observing the anatomical structure of brain, the pathology cannot be seen in T1 image. However, as shown in Fig. \ref{fig_MRI}(b), the pathology (bright area) is easily discovered from the T2 image, but the normal brain tissues have almost the same gray level with the ventricles and cerebrospinal fluid. Therefore, these two images exhibit different abilities of discernment for the involved three areas: pathology, normal brain tissues, and ventricles and cerebrospinal fluid.

\begin{figure}[!ht]
\centering
\subfigure[Lower (red) and upper (red + blue) approximations of cluster 1 (ventricles and cerebrospinal fluid)]{
\includegraphics[scale=0.6]{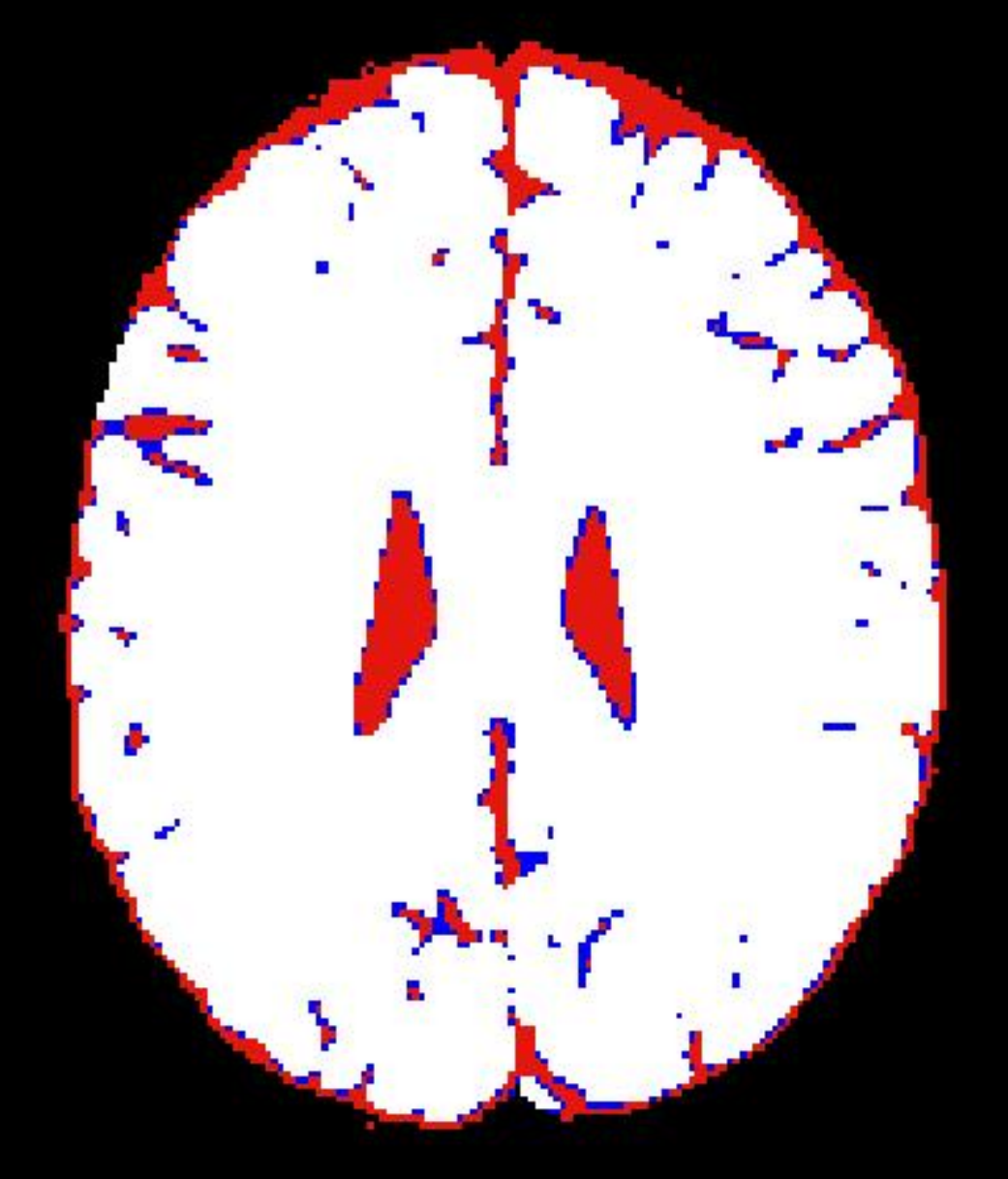}}
\subfigure[Lower (red) and upper (red + blue) approximations of cluster 2 (brain tissues)]{
\includegraphics[scale=0.6]{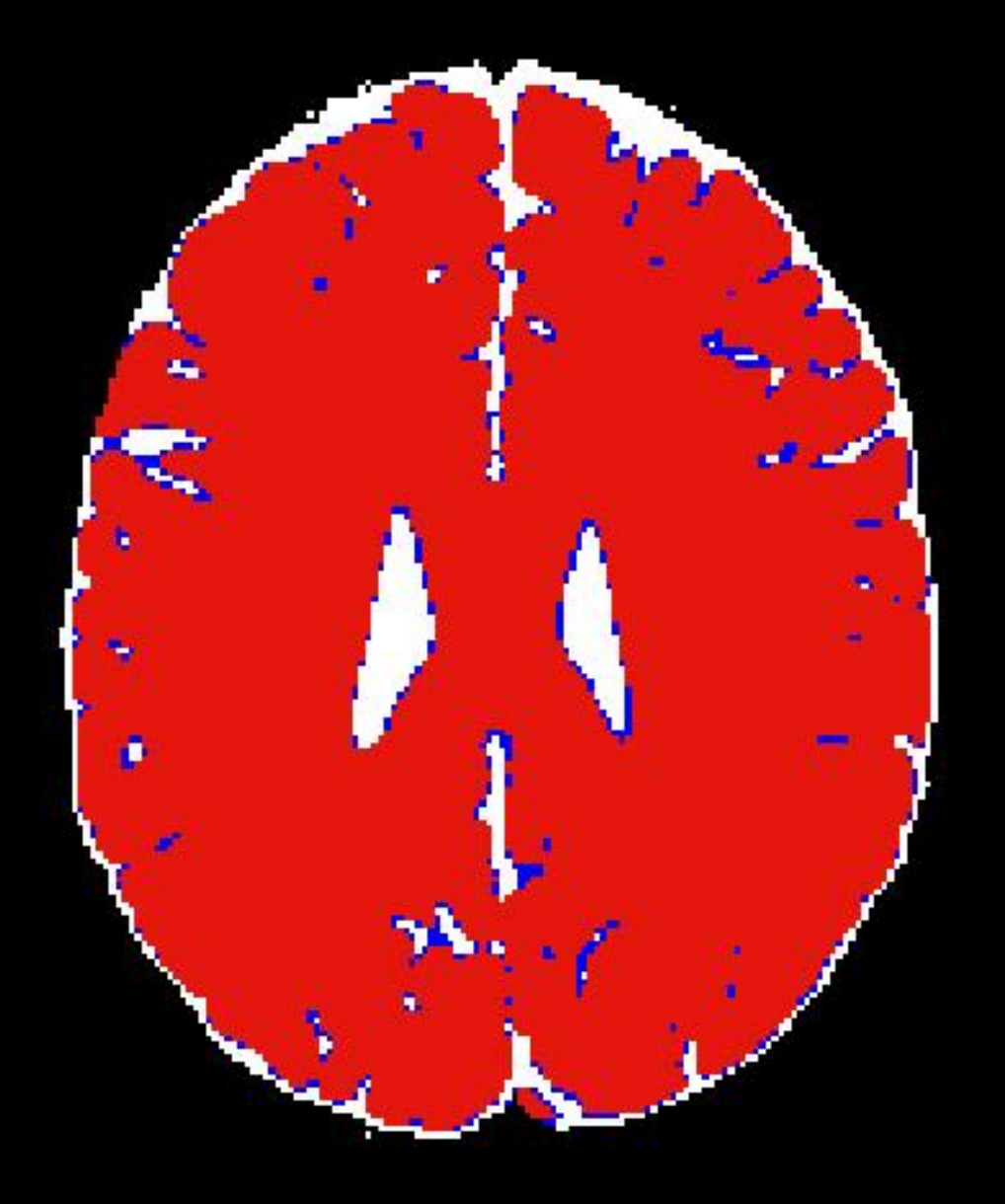}}
\caption{Segmentation results of the T1 brain image}
\label{fig_MRI_T1}
\end{figure}

\begin{figure}[!ht]
\centering
\subfigure[Lower (red) and upper (red + blue) approximations of cluster 1 (pathology)]{
\includegraphics[scale=0.6]{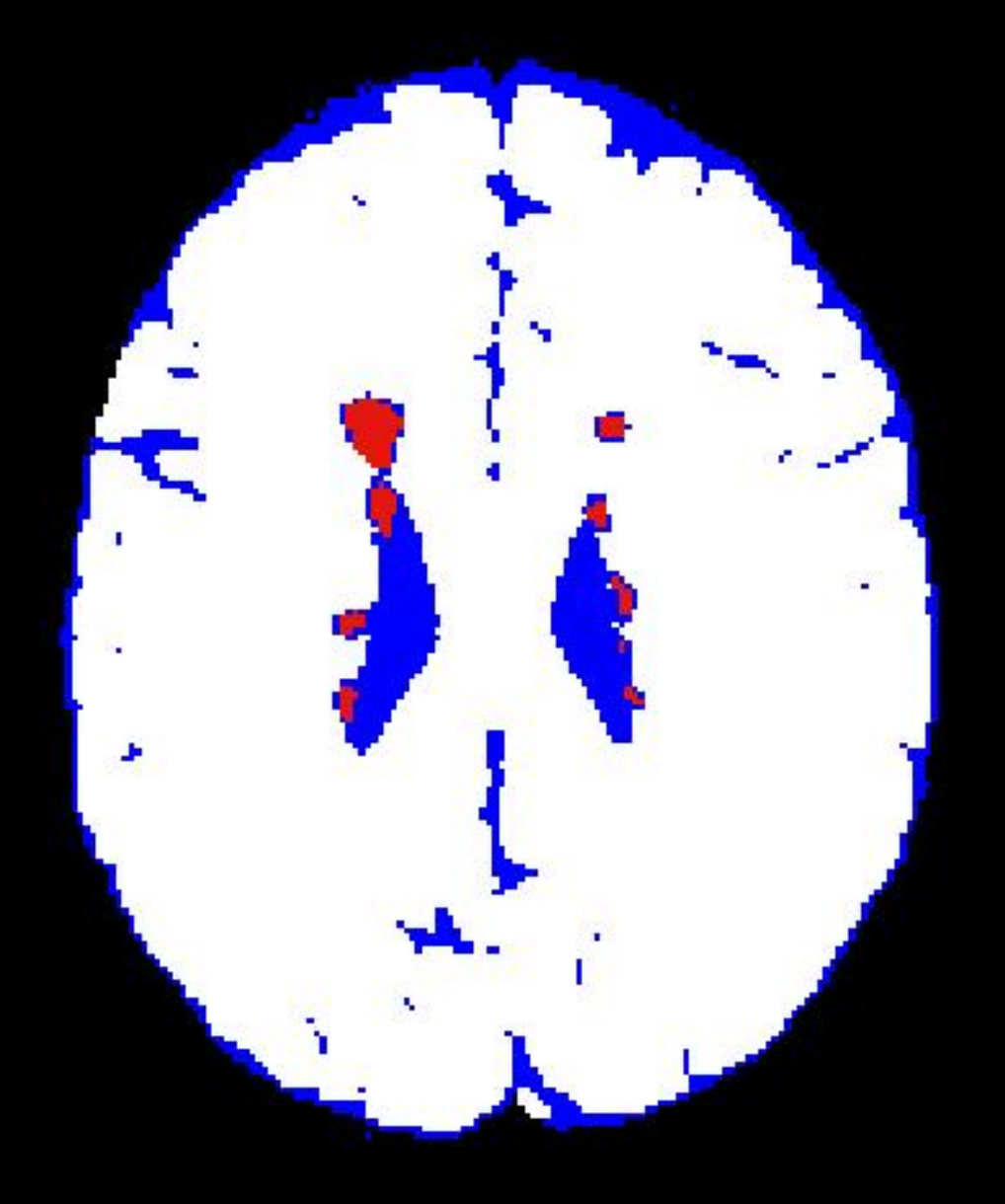}}
\subfigure[Lower (red) and upper (red + blue) approximations of cluster 2 (ventricles and cerebrospinal fluid, normal brain tissues)]{
\includegraphics[scale=0.6]{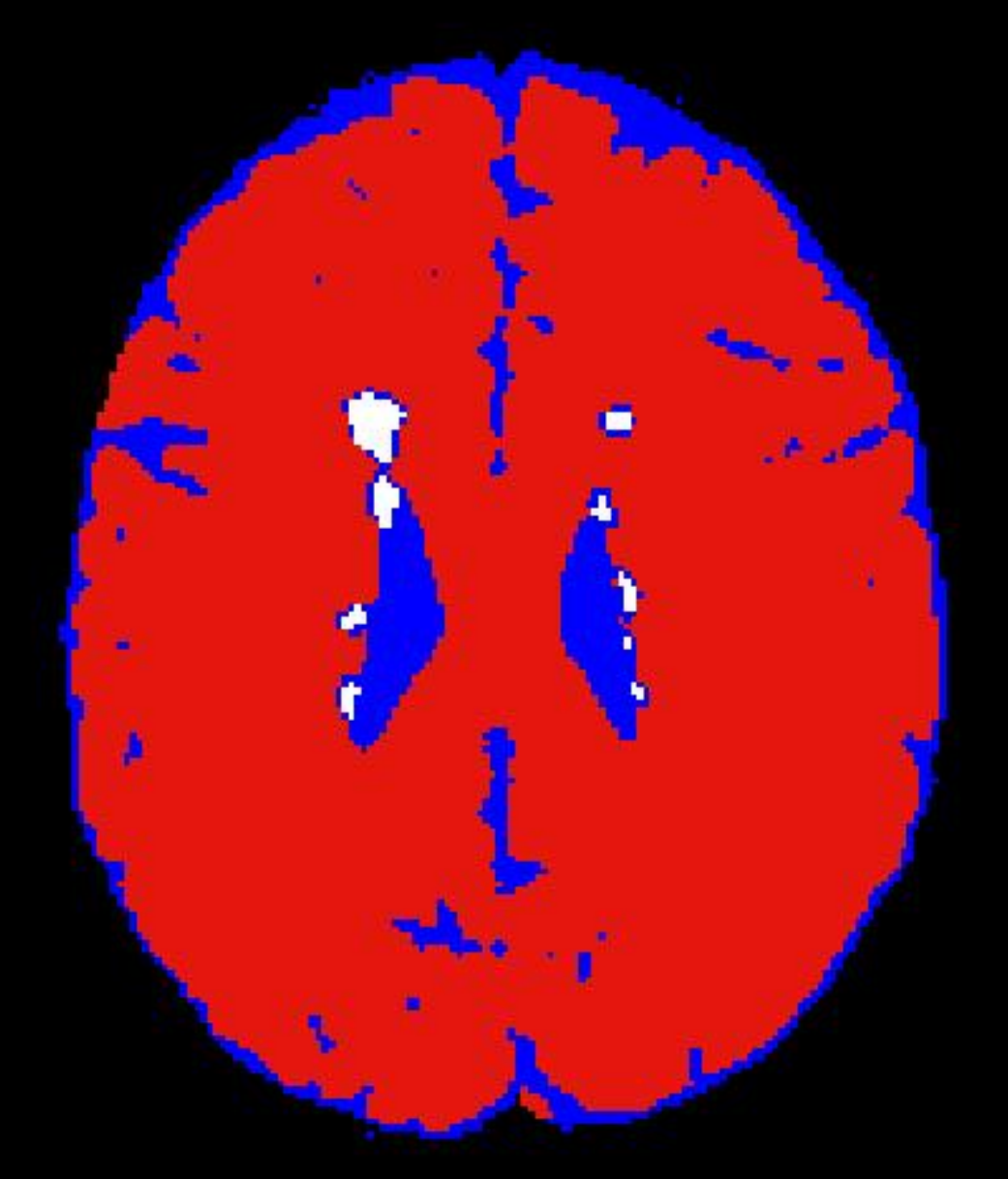}}
\caption{Segmentation results of the T2 brain image}
\label{fig_MRI_T2}
\end{figure}

The proposed EGMM was first applied on each individual image to get the evidential partition results under each modality. For each image, we run the EGMM algorithm under different numbers of clusters and calculate the corresponding values of the EBIC index. The maximum is obtained when two clusters are partitioned for both of the two images, which is consistent with our visual perception. Figs. \ref{fig_MRI_T1} and \ref{fig_MRI_T2} show the segmentation results of the T1 image and T2 image, respectively. For each cluster, the lower and upper approximations are provided, which represent the pessimistic and the optimistic decisions, respectively. As seen from Fig. \ref{fig_MRI_T1}, for T1 image, the ventricles and cerebrospinal fluid are well separated from brain tissues in terms of the lower approximations, and the blue area represents the ambiguous pixels lying at the cluster boundary. For T2 image, as shown in Fig. \ref{fig_MRI_T2}, the pathology is recovered by the lower approximation of cluster 1, and the other area (ventricles and cerebrospinal fluid as well as normal brain tissues) is recovered by the upper approximation of cluster 2. Note that a large number of ambiguous pixels are detected which take intermediate gray levels in T2 image.

\begin{figure}[!ht]
\centering
\subfigure[Lower (red) and upper (red + blue) approximations of cluster 1 (pathology)]{
\includegraphics[scale=0.6]{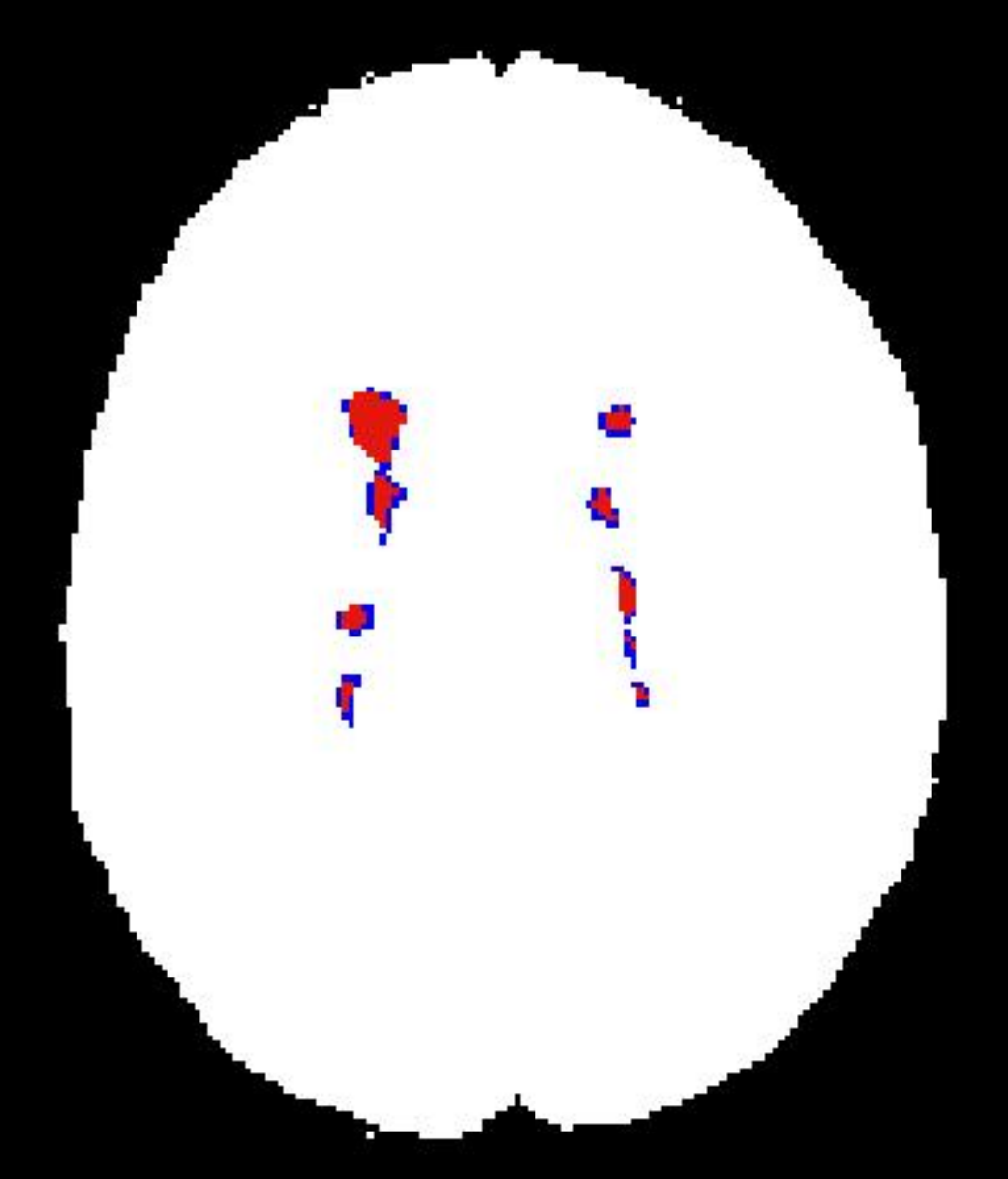}}
\subfigure[Lower (red) and upper (red + blue) approximations of cluster 2 (ventricles and cerebrospinal fluid)]{
\includegraphics[scale=0.6]{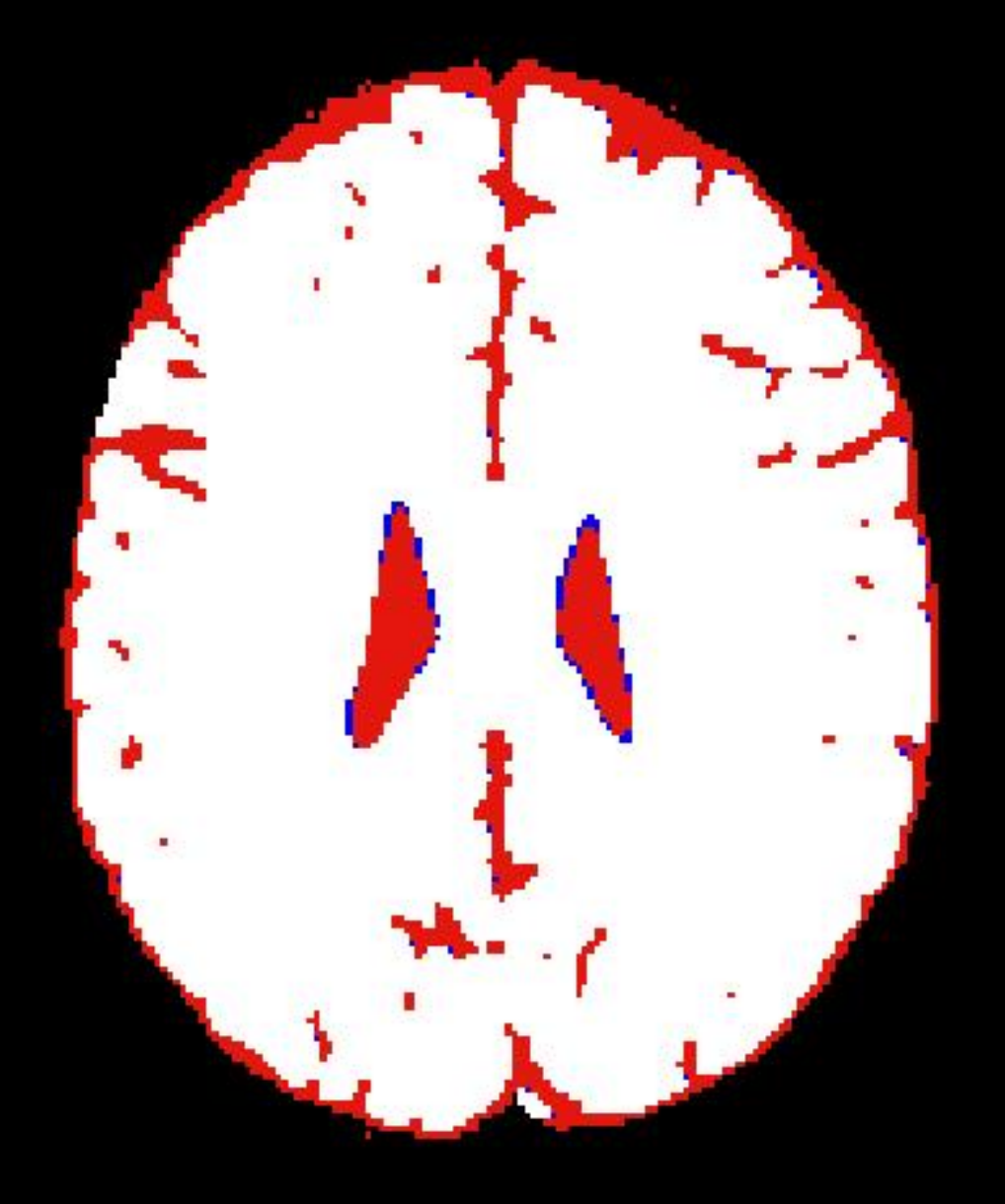}}
\subfigure[Lower (red) and upper (red + blue) approximations of cluster 3 (normal brain tissues)]{
\includegraphics[scale=0.6]{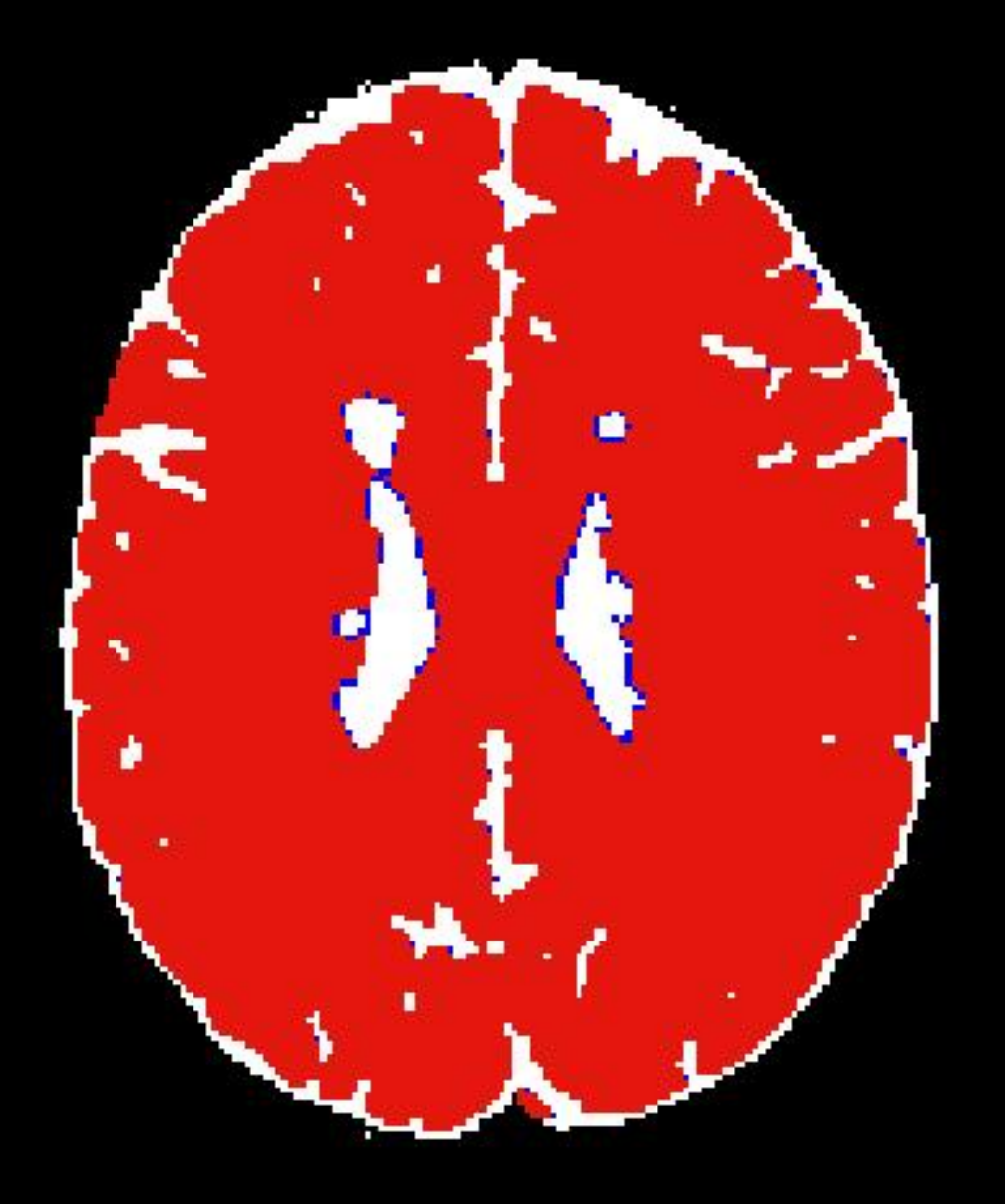}}
\caption{Segmentation results by combining the evidential partitions of T1 and T2 brain images}
\label{fig_MRI_T12}
\end{figure}

Next, we combined the above two two-cluster partitions in order to get a meaningful three-cluster partition of the brain with lower ambiguity. As the clustering results of the proposed EGMM are characterized by mass functions, for each pixel, the fusion result can be obtained by combining the mass functions derived from the two images using the classical Dempster's rule \cite{Dempster67}. Fig. \ref{fig_MRI_T12} shows the segmentation results by combining the evidential partitions of T1 and T2 brain images. It can be seen that, the three areas of interest (pathology, ventricles and cerebrospinal fluid, normal brain tissues) are well separated by the lower approximations. Besides, the number of ambiguous pixels (with blue color) are greatly reduced after fusion. This application example demonstrates the great expressive power of the evidential partition induced by the proposed EGMM. Compared with hard partition, the generated evidential partition is more informative to characterize the clustering ambiguity, which can be further reduced by later fusion process.

To demonstrate the superiority of the proposed EGMM for multi-modal brain image segmentation, we compared its performance with other representative evidential clustering algorithms (the bootGMM is not included as it cannot work on this large data). Table \ref{tab_CPU} shows the performance comparison of different algorithms on three criteria purity, NMI and ARI, from which, it can be seen that the proposed EGMM outperforms the others on all of the three criteria.

\begin{table*}[!ht]
\caption{Performance comparison of different algorithms for multi-modal brain image segmentation}
\begin{center}
\begin{tabular}{p{21mm}p{15mm}p{15mm}p{12mm}} 
\hline
Algorithm     &Purity     &NMI    &ARI \\
\hline
ECM           &0.9777     &0.8148          &0.9598  \\
BPEC          &0.9542     &0.7965          &0.9321  \\
EGMM          &\textbf{0.9844}     &\textbf{0.8323}          &\textbf{0.9721}  \\
\hline
\end{tabular}
\end{center} \label{tab_MRI}
\end{table*}
\section{Conclusions} \label{sec5}
In this paper, a new model-based approach to evidential clustering has been proposed. It is based on the notion of evidential partition, which extends those of probabilistic (or fuzzy), possibilistic, and rough ones. Different from the approximately calibrated approach in \cite{Denoeux20}, our proposal generates the evidential partition directly by searching for the maximum likelihood solution of the new EGMM via EM algorithm. In addition, a validity index is presented to determine the number of clusters automatically. The proposed EGMM is so simple that it has no open parameter and the convergence properties can also be well guaranteed. More importantly, the generated evidential partition provides a more complete description of the clustering structure than does the probabilistic (or fuzzy) partition provided by the GMM.

The proposed EGMM has been compared to some representative clustering algorithms using both synthetic and real datasets. It was shown to outperform other algorithms for a majority of datasets and rank the first in average on the three criteria clustering purity, NMI and ARI. The computation time analysis showed that the proposed EGMM runs fastest among the considered evidential clustering algorithms. Besides, the multi-modal brain image segmentation application demonstrated the great expressive power of the evidential partition induced by the proposed EGMM.

As indicated in \cite{Banfield93}, different kinds of constraints can be imposed on the covariance matrices of the GMM, which results in a total of fourteen models with different assumptions on the shape, volume and orientation of the clusters. In our work, the commonly used model with equal covariance matrix is adopted to develop the evidential clustering algorithm. It is quite interesting to further study evidential versions of the GMM with other constraints. This research direction will be explored in future work.

\appendix
\section{EM solution for the EGMM: The E-step} \label{secA}
In the E-step, we need to derive the $Q$-function, by computing the conditional expectation of observed-data log-likelihood $\log \mathcal{L}_C (\mathbf{\widetilde{\Theta}})$ given $\mathbf{X}$, using the current fit for $\widetilde{\Theta}$, i.e.,
\begin{equation}
Q(\widetilde{\Theta}, \widetilde{\Theta}^{(s)}) = \mathbb{E}_{\widetilde{\Theta}^{(s)}}\left[\log \mathcal{L}_C (\mathbf{\widetilde{\Theta}})\mid \mathbf{X}\right].
\end{equation}
By bringing the expression of the observed-data log-likelihood $\log \mathcal{L}_C (\mathbf{\widetilde{\Theta}})$ in Eq. (\ref{eq_CL}) into the above formula, we have
\begin{equation}
\begin{array}{*{20}{l}}
Q(\widetilde{\Theta}, \widetilde{\Theta}^{(s)}) = \mathbb{E}_{\widetilde{\Theta}^{(s)}}\left[\log \mathcal{L}_C (\mathbf{\widetilde{\Theta}})\mid \mathbf{X}\right]
&=& \mathbb{E}_{\widetilde{\Theta}^{(s)}}\left\{\sum \limits_{i=1} \limits^{N} \sum \limits_{j=1} \limits^{M} z_{ij} \log \left[\widetilde{\pi}_j \mathcal{N}(\mathbf{x}_i\mid \mathbf{\widetilde{\mu}}_j,\mathbf{\widetilde{\Sigma}}_j)\right] \mid \mathbf{X}\right\}\\
&=& \sum \limits_{i=1} \limits^{N} \sum \limits_{j=1}\limits^{M} \mathbb{E}_{\widetilde{\Theta}^{(s)}} \left[ z_{ij} \mid \mathbf{x}_i\right] \log \left[\widetilde{\pi}_j \mathcal{N}(\mathbf{x}_i\mid \mathbf{\widetilde{\mu}}_j,\mathbf{\widetilde{\Sigma}}_j)\right],
\end{array}
\end{equation}
with
\begin{equation}
\mathbb{E}_{\widetilde{\Theta}^{(s)}} \left[ z_{ij} \mid \mathbf{x}_i\right] = p_{\widetilde{\Theta}^{(s)}} \left[ z_{ij}=1 \mid \mathbf{x}_i\right]\\
= \frac{\widetilde{\pi}_j^{(s)} \mathcal{N}(\mathbf{x}_i \mid \mathbf{\widetilde{\mu}}_j^{(s)},\mathbf{\widetilde{\Sigma}}_j^{(s)})}{\sum_{l=1}^{M} \widetilde{\pi}_l^{(s)} \mathcal{N}(\mathbf{x}_i \mid \mathbf{\widetilde{\mu}}_l^{(s)},\mathbf{\widetilde{\Sigma}}_l^{(s)})}\\
\triangleq m_{ij}, ~~i=1,\ldots,N, ~j=1,\ldots,M.
\end{equation}
from which we obtain Eqs. (\ref{eq_Qfunc}) and (\ref{eq_EMemb}) of the E-step.
\section{EM solution for the EGMM: The M-step} \label{secB}
In the M-step, we need to maximize the $Q$-function $Q(\widetilde{\Theta}, \widetilde{\Theta}^{(s)})$ derived in the E-step with respect to the involved parameters: the mixing probabilities of the $M$ components $\{\widetilde{\pi}_j\}_{j=1}^{M}$, the mean vectors of the $C$ single-clusters $\{\mu_k\}_{k=1}^{C}$ and the common covariance matrix $\mathbf{\Sigma}$. To find a local maxima, we compute the derivatives of $Q(\widetilde{\Theta}, \widetilde{\Theta}^{(s)})$ with respect to $\{\widetilde{\pi}_j\}_{j=1}^{M}$, $\{\mu_k\}_{k=1}^{C}$, and $\mathbf{\Sigma}$, respectively.

First, we compute the derivative of $Q(\widetilde{\Theta}, \widetilde{\Theta}^{(s)})$ with respect to the mixing probabilities $\{\widetilde{\pi}_j\}_{j=1}^{M}$. Notice that the values of $\{\widetilde{\pi}_j\}_{j=1}^{M}$ are constrained to be positive and sum to one. This constraint can be handled using a Lagrange multiplier and maximizing the following quantity:
\begin{equation}
\mathcal{F} \left(\{\widetilde{\pi}_j\}_{j=1}^{M},\lambda\right) = Q(\widetilde{\Theta}, \widetilde{\Theta}^{(s)}) - \lambda \left(\sum \limits_{j=1}\limits^{M}{\widetilde{\pi}_j} -1\right) .
\end{equation}
By differentiating $\mathcal{F}$ with respect to $\widetilde{\pi}_j$ and $\lambda$ and setting the derivatives to zero, we obtain:
\begin{equation}
\frac{\partial \mathcal{F}}{\partial \widetilde{\pi}_j} = \frac{1}{\widetilde{\pi}_j} \sum \limits_{i=1} \limits^{N} m_{ij}^{(s)} - \lambda = 0, ~~j=1,\ldots,M, \label{eq_Dpi}
\end{equation}
\begin{equation}
\frac{\partial \mathcal{F}}{\partial \lambda} = 1 - \sum \limits_{j=1}\limits^{M}{\widetilde{\pi}_j} = 0. \label{eq_Dlam}
\end{equation}
From Eq. (\ref{eq_Dpi}) we have
\begin{equation}
\widetilde{\pi}_j = \frac{1}{\lambda} \sum \limits_{i=1} \limits^{N} m_{ij}^{(s)}, ~~j=1,\ldots,M. \label{eq_pi}
\end{equation}
By Bringing the above expression of $\widetilde{\pi}_j $ into Eq. (\ref{eq_Dlam}), we have
\begin{equation}
1 - \sum \limits_{j=1}\limits^{M}\left({\frac{1}{\lambda} \sum \limits_{i=1} \limits^{N} m_{ij}^{(s)}}\right) = 0 ~~\Leftrightarrow~~ 1 - {\frac{1}{\lambda} \sum \limits_{i=1} \limits^{N} \sum \limits_{j=1}\limits^{M}  m_{ij}^{(s)}} = 0
\end{equation}
Notice that the sum of the evidential membership for each object is one, i.e., $\sum _{j=1}^{M}  m_{ij}^{(s)} = 1$, $i=1,\ldots,N$. We thus have $\lambda = N$. Then, returning in Eq. (\ref{eq_pi}), we obtain the updated mixing probabilities $\widetilde{\pi}_j$ as
\begin{equation}
\widetilde{\pi}_j = \frac{1}{N} \sum \limits_{i=1} \limits^{N} m_{ij}^{(s)}, ~~j=1,\ldots,M.
\end{equation}

Second, we compute the derivative of $Q(\widetilde{\Theta}, \widetilde{\Theta}^{(s)})$ with respect to the mean vectors of single-clusters $\{\mu_k\}_{k=1}^{C}$:
\begin{equation}
\frac{\partial Q}{\partial \mu_k} = \sum \limits_{i=1} \limits^{N} \sum \limits_{j=1} \limits^{M} m_{ij}^{(s)} \frac{\partial \mathcal{N}(\mathbf{x}_i\mid \mathbf{\widetilde{\mu}}_j,\mathbf{\widetilde{\Sigma}}_j)}{\partial \mu_k}
= \sum \limits_{i=1} \limits^{N} \sum \limits_{j=1} \limits^{M} m_{ij}^{(s)} |A_j|^{-1}I_{kj}{\mathbf{\widetilde{\Sigma}}_j}^{-1}(\mathbf{x}_i-\mathbf{\widetilde{\mu}}_j)
= \sum \limits_{i=1} \limits^{N} \sum \limits_{j=1} \limits^{M} m_{ij}^{(s)} |A_j|^{-1}I_{kj}{\mathbf{\widetilde{\Sigma}}_j}^{-1}(\mathbf{x}_i-\mathbf{\widetilde{\mu}}_j),
~k=1,\ldots,C.
\end{equation}

By bringing the expressions of the mean vectors $\mathbf{\widetilde{\mu}}_j$ and the covariance matrices $\mathbf{\widetilde{\Sigma}}_j$ in Eqs. (\ref{eq_ConsMean}) and (\ref{eq_ConsCov}) into the above formula, we have
\begin{equation}
\frac{\partial Q}{\partial \mu_k} =  \sum \limits_{i=1} \limits^{N} \sum \limits_{j=1} \limits^{M} m_{ij}^{(s)} |A_j|^{-1}I_{kj}{\mathbf{\Sigma}}^{-1}(\mathbf{x}_i-|A_j|^{-1} \sum \limits_{l=1} \limits^{C} I_{lj} \mathbf{\mu}_l), ~~k=1,\ldots,C.
\end{equation}
Setting these derivatives to zero and doing simple transformation, we obtain
\begin{equation}
\sum \limits_{i=1} \limits^{N} \mathbf{x}_i  \sum \limits_{j=1} \limits^{M} |A_j|^{-1} m_{ij}^{(s)} I_{kj} = \sum \limits_{l=1} \limits^{C} \mathbf{\mu}_l \sum \limits_{i=1} \limits^{N} \sum \limits_{j=1} \limits^{M} |A_j|^{-2} m_{ij}^{(s)} I_{kj} I_{lj}, ~~k=1,\ldots,C.
\end{equation}
Using the notations in Eqs. (\ref{eq_H}) and (\ref{eq_B}), we obtain the updated mean vectors of single-clusters $\mathbf{\Xi} = [\mathbf{\mu}_1;\ldots; \mathbf{\mu}_C]$ as
\begin{equation}
\mathbf{\Xi} = \mathbf{H}^{-1}\mathbf{B}.
\end{equation}

At last, we compute the derivative of $Q(\widetilde{\Theta}, \widetilde{\Theta}^{(s)})$ with respect to the common covariance matrix $\mathbf{\Sigma}$:
\begin{equation}
\frac{\partial Q}{\partial \mathbf{\Sigma}} = \sum \limits_{i=1} \limits^{N} \sum \limits_{j=1} \limits^{M} m_{ij}^{(s)} \frac{\partial \mathcal{N}(\mathbf{x}_i \mid \mathbf{\widetilde{\mu}}_j,\mathbf{\widetilde{\Sigma}}_j)}{\partial \mathbf{\Sigma}}\\
= \sum \limits_{i=1} \limits^{N} \sum \limits_{j=1} \limits^{M} m_{ij}^{(s)} (-\frac{1}{2}) \mathbf{\Sigma}^{-1} [1-(\mathbf{x}_i-\mathbf{\widetilde{\mu}}_j)(\mathbf{x}_i-\mathbf{\widetilde{\mu}}_j)^{T}] \mathbf{\Sigma}^{-1}. \\
\end{equation}
Setting this derivative to zero and doing simple transformation, we obtain
\begin{equation}
\left(\sum \limits_{i=1} \limits^{N} \sum \limits_{j=1} \limits^{M} m_{ij}^{(s)}\right) \mathbf{\Sigma} = \sum \limits_{i=1} \limits^{N} \sum \limits_{j=1} \limits^{M} m_{ij}^{(s)} (\mathbf{x}_i-\mathbf{\widetilde{\mu}}_j)(\mathbf{x}_i-\mathbf{\widetilde{\mu}}_j)^{T}.
\end{equation}
Notice that the sum of the evidential membership for each object is one, i.e., $\sum _{j=1}^{M}  m_{ij}^{(s)} = 1$, $i=1,\ldots,N$. We thus obtain the updated common covariance matrix $\mathbf{\Sigma}$ as
\begin{equation}
\mathbf{\Sigma} = \frac{1}{N} \sum _{i=1}^{N} \sum _{j=1}^{M} m_{ij}^{(s)} (\mathbf{x}_i - \mathbf{\widetilde{\mu}}_j)(\mathbf{x}_i - \mathbf{\widetilde{\mu}}_j)^T.
\end{equation}

\section*{Acknowledgments}
This work was funded by the National Natural Science Foundation of China (Grant Nos. 62171386, 61801386 and 61790552), the Key Research and Development Program in Shaanxi Province of China (Grant No. 2022GY-081), and the China Postdoctoral Science Foundation (Grant Nos. 2020T130538 and 2019M653743).

\section*{References}
\bibliographystyle{elsarticle-num}
\bibliography{MyBib}






\end{document}